\documentclass[11pt]{article}
\usepackage{fullpage}
\usepackage{titling}
\usepackage{changepage}
\usepackage{url}            
\usepackage{graphicx}
\usepackage{natbib}
\usepackage{xspace}
\usepackage{booktabs} 
\usepackage{caption}
\usepackage{subcaption}
\usepackage{xcolor}
\usepackage{tcolorbox}
\usepackage{enumerate}
\usepackage{enumitem}
\usepackage{amsfonts}
\usepackage{multirow}
\usepackage{lipsum}
\usepackage{makecell}
\usepackage{tikz}
\usepackage{geometry}

\definecolor{skyblue}{RGB}{65,105,225}

\usepackage[
pagebackref=true,
citecolor=skyblue,
linkcolor=red,
urlcolor=skyblue,
menucolor=black,
letterpaper=true,
breaklinks=true,
bookmarks=true,
colorlinks
]{hyperref}

\bibpunct{[}{]}{,}{n}{}{;}
\newcommand{\modelname}{Tarsier2\xspace}

\title{\modelname: Advancing Large Vision-Language Models from Detailed Video Description to Comprehensive Video Understanding
}

\author{ 
Liping Yuan$^*$ \qquad Jiawei Wang$^*$ \qquad Haomiao Sun$^{*}$ \qquad Yuchen Zhang$^{*}$ \qquad Yuan Lin$^{\dag}$ \\\\
ByteDance Research\\
{
\small
\texttt{
\{yuanliping.0o0,wangjiawei.424,sunhaomiao,zhangyuchen.zyc,linyuan.0\}@bytedance.com
}
}
\small\\Project Site: \url{https://github.com/bytedance/tarsier}
}

\date{}

\begin{document}
\setlength{\droptitle}{-8em} 
\begin{adjustwidth}{-0.1\textwidth}{-0.1\textwidth}
\begin{center}
    \begin{minipage}{1.2\textwidth} 
        \centering
        \maketitle 
    \vspace*{-1.4cm}
    \begin{abstract}
    \noindent We introduce \modelname, a state-of-the-art large vision-language model (LVLM) designed for generating detailed and accurate video descriptions, while also exhibiting superior general video understanding capabilities. \modelname achieves significant advancements through three key upgrades: (1) Scaling pre-training data from 11M to 40M video-text pairs, enriching both volume and diversity; (2) Performing fine-grained temporal alignment during supervised fine-tuning; (3) Using model-based sampling to automatically construct preference data and applying DPO training for optimization. Extensive experiments show that \modelname-7B consistently outperforms leading proprietary models, including GPT-4o and Gemini 1.5 Pro, in detailed video description tasks. On the DREAM-1K benchmark, \modelname-7B improves F1 by 2.8\% over GPT-4o and 5.8\% over Gemini-1.5-Pro. In human side-by-side evaluations, \modelname-7B shows a +8.6\% performance advantage over GPT-4o and +24.9\% over Gemini-1.5-Pro. \modelname-7B also sets new state-of-the-art results across 15 public benchmarks, spanning tasks such as video question-answering, video grounding, hallucination test, and embodied question-answering, demonstrating its versatility as a robust generalist vision-language model.
    \end{abstract}
    \end{minipage}
\end{center}
\end{adjustwidth}

\begin{figure}[h!]
    \centering
    \resizebox{0.85\textwidth}{!}{
    \begin{minipage}{0.6\textwidth}
        \centering
        \includegraphics[width=\textwidth]{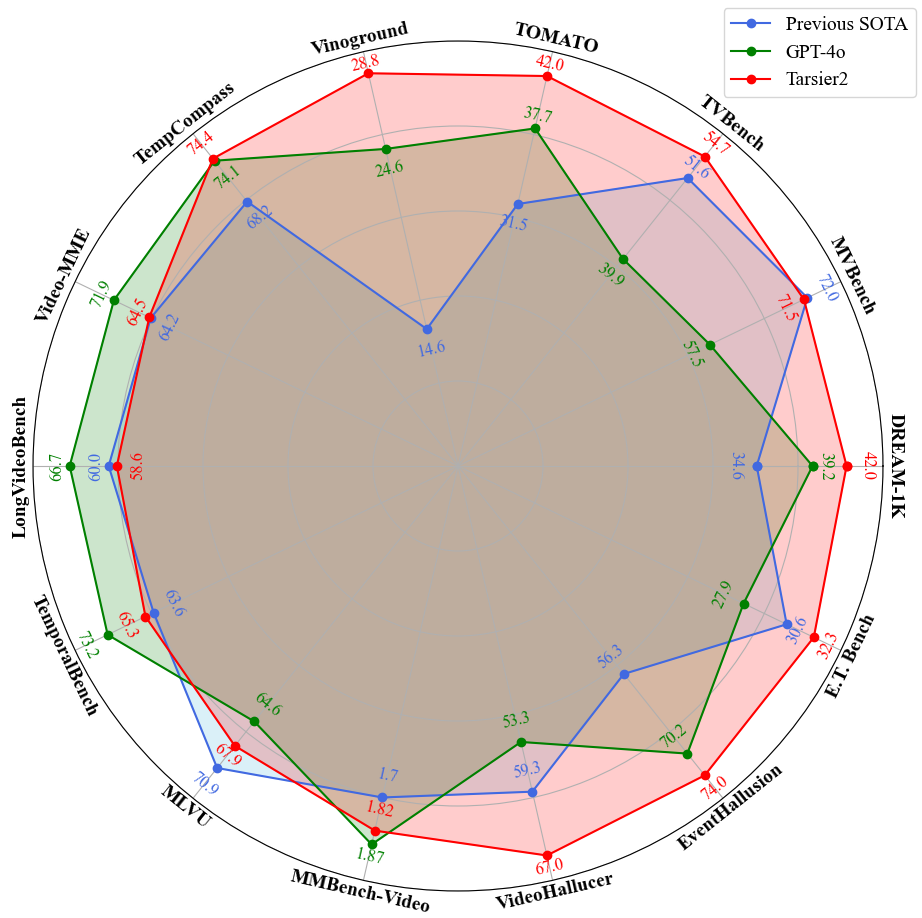} 
    \end{minipage}\hfill
    \quad\begin{minipage}{0.38\textwidth}
        \centering
        \resizebox{\textwidth}{!}{
        \begin{tabular}{lc}
            \toprule
             \textbf{Benchmark} & \textbf{Previous SOTA} \\
            \midrule
            DREAM-1K\cite{wang2024tarsierrecipestrainingevaluating} & Tarsier-7B\cite{wang2024tarsierrecipestrainingevaluating} \\
            MVBench\cite{li2024mvbench} &  InternVL2.5-8B\cite{chen2024expanding}\\
            TVBench\cite{cores2024tvbench} & IXC-2.5 7B\cite{zhang2024internlmxcomposer} \\
            TOMATO\cite{shangguan2024tomato} & Qwen2-VL-7B\cite{qwen2vl} \\
            Vinoground\cite{zhang2024vinoground} & LLaVA-OV-7B\cite{li2024llavanext} \\
            TempCompass\cite{liu2024tempcompass} & Qwen2-VL-7B\cite{qwen2vl} \\
            Video-MME\cite{fu2024video} & NVILA-7B\cite{liu2024nvila} \\
            LongVideoBench\cite{wu2024longvideobench} & Apollo-7B\cite{zohar2024apolloexplorationvideounderstanding} \\
            TemporalBench\cite{cai2024temporalbench} & LLaVA-Video-7B\cite{zhang2024video} \\
            MLVU\cite{zhou2024mlvu} & InternVL2.5-8B\cite{chen2024expanding} \\
            MMBench-Video\cite{fang2024mmbench} & MiniCPM-V-2.6 \cite{yao2024minicpm} \\
            VideoHallucer\cite{wang2024videohallucer} & Qwen2-VL-7B\cite{qwen2vl} \\
            EventHallusion\cite{zhang2024eventhallusion} & Tarsier-7B\cite{wang2024tarsierrecipestrainingevaluating} \\
            E.T. Bench\cite{liu2024etbench} & E.T. Chat\cite{liu2024etbench} \\
            \bottomrule
        \end{tabular}
        }
    \end{minipage}
    }
    \caption{\small{Performance comparison of Tarsier2 with previous SOTA models at 7B-scale and GPT-4o. We report the overall average scores for benchmarks with multiple subtasks/metrics.}}
    \label{fig:radar}
\end{figure}
\renewcommand{\thefootnote}{}
\footnotetext[1]{$*$Equally contributed. $\dag$Corresponding author.}
\renewcommand{\thefootnote}{\arabic{footnote}}



\newpage
    {\hypersetup{linkcolor=skyblue}
    \tableofcontents}
\newpage


\section{Introduction}

With the rapid advancements in large vision-language models (LVLM)~\cite{chen2024far,li2023videochat,lin2023video,lin2024vila,wang2024tarsierrecipestrainingevaluating,qwen2vl}, significant progress has also been made in video understanding. Leading proprietary models, such as GPT-4o~\cite{gpt4o} and Gemini-1.5-Pro~\cite{geminiteam2024gemini15unlockingmultimodal}, have achieved state-of-the-art (SOTA) performance across a variety of video understanding tasks. Additionally, several open-source models~\cite{lin2023video,xu2024pllava,cheng2024videollama2,li2024llava,chen2024expanding,li2024llavanext,cheng2024videollama2} also demonstrate strong performance on several video understanding benchmarks~\cite{cores2024tvbench,li2024mvbench,liu2024etbench,wang2024videohallucer,zhou2024mlvu}, although they still lag behind proprietary models, particularly in complex, open-ended generation tasks. Despite these advancements, current models remain behind human-level video understanding~\cite{mangalam2023egoschema,patraucean2024perception,chen2023autoeval}, mainly due to persistent challenges such as accurately perceiving temporal dynamics, spatial-temporal reasoning, and model hallucinations.

\begin{figure}[p]
    \vspace*{-2cm}
    \makebox[\linewidth]{
    \centering
    \begin{tikzpicture}
        \node [anchor=south west,inner sep=0] (image) at (0,0) {\includegraphics[width=\textwidth]{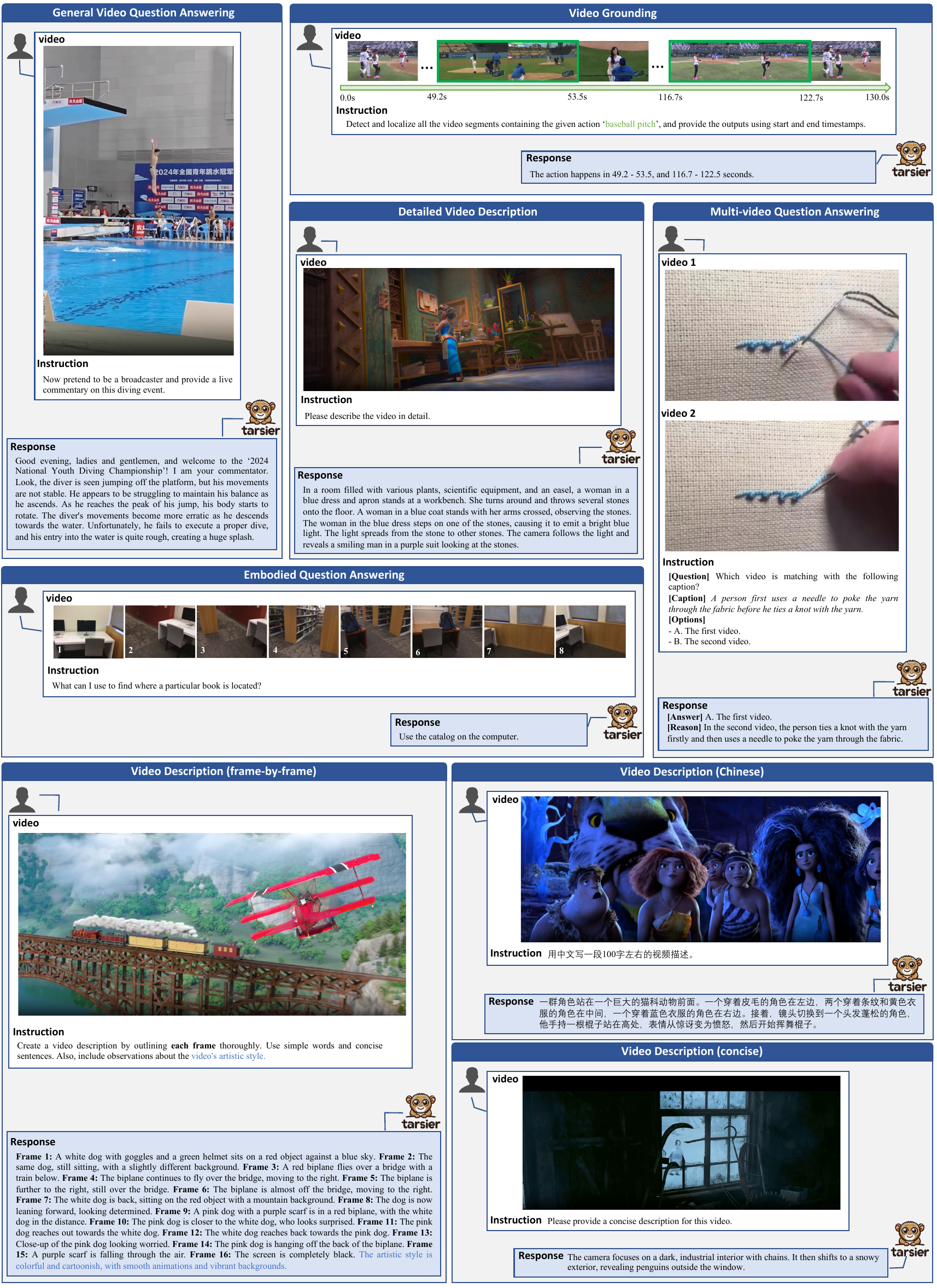}};
        \begin{scope}[x={(image.south east)},y={(image.north west)}]
            \node[anchor=north west] at (0.2,0.76) {\href{https://dream-videos.s3.us-east-1.amazonaws.com/26327764.mp4}{\includegraphics[width=0.04\textwidth]{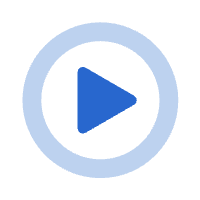}}};
        \end{scope}
        \begin{scope}[x={(image.south east)},y={(image.north west)}]
            \node[anchor=north west] at (0.608,0.733) {\href{https://dream-videos.s3.amazonaws.com/149.mp4}{\includegraphics[width=0.04\textwidth]{figs/play.png}}};
        \end{scope}
        \begin{scope}[x={(image.south east)},y={(image.north west)}]
            \node[anchor=north west] at (0.909,0.727) {\href{https://dream-videos.s3.us-east-1.amazonaws.com/145_pos.mp4}{\includegraphics[width=0.04\textwidth]{figs/play.png}}};
        \end{scope}
        \begin{scope}[x={(image.south east)},y={(image.north west)}]
            \node[anchor=north west] at (0.909,0.61) {\href{https://dream-videos.s3.us-east-1.amazonaws.com/145_neg.mp4}{\includegraphics[width=0.04\textwidth]{figs/play.png}}};
        \end{scope}
        \begin{scope}[x={(image.south east)},y={(image.north west)}]
            \node[anchor=north west] at (0.382,0.25) {\href{https://dream-videos.s3.us-east-1.amazonaws.com/30.mp4}{\includegraphics[width=0.04\textwidth]{figs/play.png}}};
        \end{scope}
        \begin{scope}[x={(image.south east)},y={(image.north west)}]
            \node[anchor=north west] at (0.888,0.305) {\href{https://dream-videos.s3.us-east-1.amazonaws.com/38.mp4}{\includegraphics[width=0.04\textwidth]{figs/play.png}}};
        \end{scope}
        \begin{scope}[x={(image.south east)},y={(image.north west)}]
            \node[anchor=north west] at (0.849,0.098) {\href{https://dream-videos.s3.us-east-1.amazonaws.com/9.mp4}{\includegraphics[width=0.04\textwidth]{figs/play.png}}};
        \end{scope}
    \end{tikzpicture}
    }
    \caption{Overview of \modelname capabilities. Based on its strong ability for detailed video description, \modelname excels in a variety of video-centric tasks. Click the play buttons to view the videos.}
    
    \label{fig:example}
\end{figure}

In this paper, we introduce \modelname, a 7B-parameter LVLM model that can outperform both GPT-4o and Gemini-1.5-Pro in generating detailed video descriptions, a fundamental challenge in video understanding. Beyond video description generation, \modelname also achieves SOTA performance across various video question-answering (VQA) benchmarks at the same model size, surpassing or closely matching the performance of proprietary models on these VQA benchmarks. Figure~\ref{fig:radar} provides a comprehensive comparison between \modelname, GPT-4o and previous SOTA results for open-source LVLMs with the same scale. Figure~\ref{fig:example} presents examples illustrating \modelname's video understanding capability across different tasks.

\modelname employs a simple model architecture consisting of a vision encoder, a vision adaptor, and a large language model (LLM). We meticulously design a three-stage training procedure: pre-training, supervised fine-tuning (SFT), and reinforcement learning (RL). In comparison with Tarsier~\cite{wang2024tarsierrecipestrainingevaluating}, \modelname features several key improvements that significantly enhance its performance:

\begin{itemize}
\item We scale up the pre-training dataset from 11 million to 40 million video-text pairs, addressing the challenge posed by the scarcity of high-quality video-text data. To achieve this, we implement meticulous filtering and sourcing. Specifically, we collect 11 million commentary videos, featuring explanations and analyses of movies and TV shows, providing rich contextual information to greatly enhance video understanding. Our experiments confirm that increasing the volume of pre-training data consistently improves model performance.

\item We construct a video description dataset containing 150K instances, each including a detailed video description along with the specific frames corresponding to each event described. During the SFT stage, we involve this dataset to provide the model with supervision on temporal fine-grained alignment. Experimental results show that, compared with traditional video-caption alignment training, this approach significantly improves accuracy in video description and reduces the hallucinations.

\item To further enhance model performance, we use the model to generate samples that automatically construct preference data for DPO training~\cite{rafailov2024direct}. To ensure high-quality preference data, we propose two methods: a negative sampling technique that uses corrupted videos to generate negative samples for preference pairs, and a preference data filtering method that employs AutoDQ~\cite{wang2024tarsierrecipestrainingevaluating} to automatically filter out pairs with minimal differences. Our experiments show that DPO training on these automatically generated preference data leads to continued performance improvements over the SFT stage.
\end{itemize}

We conduct extensive experiments to evaluate \modelname against both proprietary and open-source LVLMs. For video description, \modelname outperforms all other models, surpassing both proprietary and open-source LVLMs in evaluations on DREAM-1K~\cite{wang2024tarsierrecipestrainingevaluating} and E.T. Bench-Captioning~\cite{liu2024etbench}. In human side-by-side evaluations, \modelname-7B shows a +7.8\% improvement over GPT-4o and a +12.3\% advantage over Gemini-1.5-Pro. It also significantly outperforms the leading open-source model, Tarsier-34B, with a +51.4\% advantage. Furthermore, \modelname-7B proves to be a versatile generalist model, setting new SOTA results on public benchmarks for video question-answering~\cite{cores2024tvbench,shangguan2024tomato,zhang2024vinoground}, hallucination test~\cite{zhang2024eventhallusion}, video grounding~\cite{liu2024etbench} and embodied QA~\cite{robovqa2023arxiv}. Finally, we present extensive ablation studies to identify the key factors contributing to the model’s strong performance. We also release a recaptioning dataset, \href{https://huggingface.co/datasets/omni-research/Tarsier2-Recap-585K}{Tarsier2-Recap-585K}, and demonstrate its effectiveness in enhancing the capabilities of existing LVLMs for video description and general video understanding.

\section{Related Work}
\paragraph{Video-LLMs} 

Recently, research on Video LLMs has surged~\cite{li2023videochat,maaz2023video,luo2023valley,zhang2023videollama,lin2023video,ataallah2024minigpt4,wang2024elysium,xu2024pllava,li2024llava,lin2024vila,zhang2024video,qwen2vl,li2024aria,dubey2024llama,agrawal2024pixtral,liu2024oryx,chen2024expanding,zohar2024apolloexplorationvideounderstanding}, with efforts focusing on model architectures and video-text data collection. On the architecture side, current studies emphasize visual representation \cite{xu2024pllava,qwen2vl,zohar2024apolloexplorationvideounderstanding}, visual token resampling \cite{xu2024pllava,chen2024expanding,xu2024slowfast,li2025llama}, and the integration of Vision Transformers (ViT) with LLMs \cite{qwen2vl,li2023blip,liu2024kangaroo,bai2023qwen}. \modelname adopts a simple architecture composed of a visual encoder, a visual adaptor, and an LLM. Despite its simplicity, we demonstrate that a meticulously designed training strategy enables \modelname to achieve strong video understanding capabilities.

In terms of video-text data, while many efforts aim to collect datasets for training Video LLMs, their quantity and quality remain limited. For example, LLaVA-Video \cite{zhang2024video} is trained on just 1.3 million video-text pairs, and several open-source models, such as InternVL2.5 \cite{chen2024expanding}, Aria \cite{li2024aria}, and VILA-1.5 \cite{lin2024vila}, are trained on fewer than 5 million pairs. Although larger datasets like HowTo100M \cite{miech2019howto100m}, HD-VILA \cite{xue2022advancing}, Panda-70M \cite{chen2024panda}, and InternVid-10M \cite{wang2023internvid} exist, they either cover limited domains or contain overly simplistic or low-quality text. Furthermore, some studies do not disclose the volume of video data used \cite{qwen2vl,zohar2024apolloexplorationvideounderstanding,dubey2024llama,li2024aria}.

To address these challenges, our work focuses on improving the quantity and quality of video-text data. We newly collected 20 million video-text pairs, spanning a wide range of video genres. In total, 40 million pairs are used in the final pre-training stage. Additionally, we annotated 150K fine-grained video descriptions for the SFT stage.

\paragraph{Video Description} 

Video description, a foundational task in video understanding, has long been a central focus of research. Early work~\cite{xu2023mplug,yan2022videococa,chen2023vast} typically involved pre-training video-language models and fine-tuning them on datasets such as MSVD\cite{chen-dolan-2011-collecting}, MSR-VTT\cite{xu2016msr}, and VATEX\cite{wang2019vatex}, which provide single-sentence video summaries.

Recent advancements in LVLMs have improved video description, enabling more detailed outputs beyond simple summarization. However, generating comprehensive video descriptions presents challenges beyond model architecture. While multi-frame processing and temporal modeling are crucial, large-scale and rich annotated <video, description> datasets are equally important. Existing alignment datasets, such as HD-VILA~\cite{xue2022advancing} and HoTo100M~\cite{miech2019howto100m}, provide concise descriptions, limiting detailed video understanding. To address this, datasets such as ShareGPT4Video\cite{chen2024sharegpt4video} uses a pipeline where LVLMs (e.g., GPT-V\cite{gpt4v}) annotate frames, and LLMs (e.g., GPT-4\cite{gpt4}) aggregate them. This improves detail but often leads to verbosity and hallucinations. Recent works~\cite{zhang2024video,tang2024enhancing} uses proprietary Video-LLMs, such as GPT-4o\cite{gpt4o} and Gemini-1.5\cite{geminiteam2024gemini15unlockingmultimodal}, for annotation, but their high cost limits application to smaller datasets.

For \modelname, we collect a large dataset of video-text pairs. In particular, we automatically build meaningful video-text pairs from online commentary videos. These commentaries include both low-level (atomic actions) and high-level (plot) visual elements, enhancing the model’s understanding across various granularity. In addition to data collection, \modelname\ also uses a meticulously designed three-stage training process, where DPO training after SFT further refines description accuracy and detail.

\section{Approach}\label{sec:approach}

We initialized Tarsier with Qwen2-VL\cite{qwen2vl} weights and employed a three-stage training strategy. First, we pre-trained \modelname on 40 million large-scale video-text pairs. Next, we fine-tuned the model on moderate-sized, curated, human-annotated datasets in two phases: one targeting video descriptions with fine-grained grounding and the other focusing on natural, instruction-following video descriptions. Finally, we applied Direct Preference Optimization\cite{rafailov2024direct} using automatically generated preference data to further enhance the quality of the video descriptions. The training process is detailed below; for a comprehensive list of hyper-parameters, please refer to Appendix~\ref{sec:training-hyperparameters}.

\subsection{Pre-training}

\begin{figure}[t]
    \centering
    \includegraphics[width=0.6\textwidth]{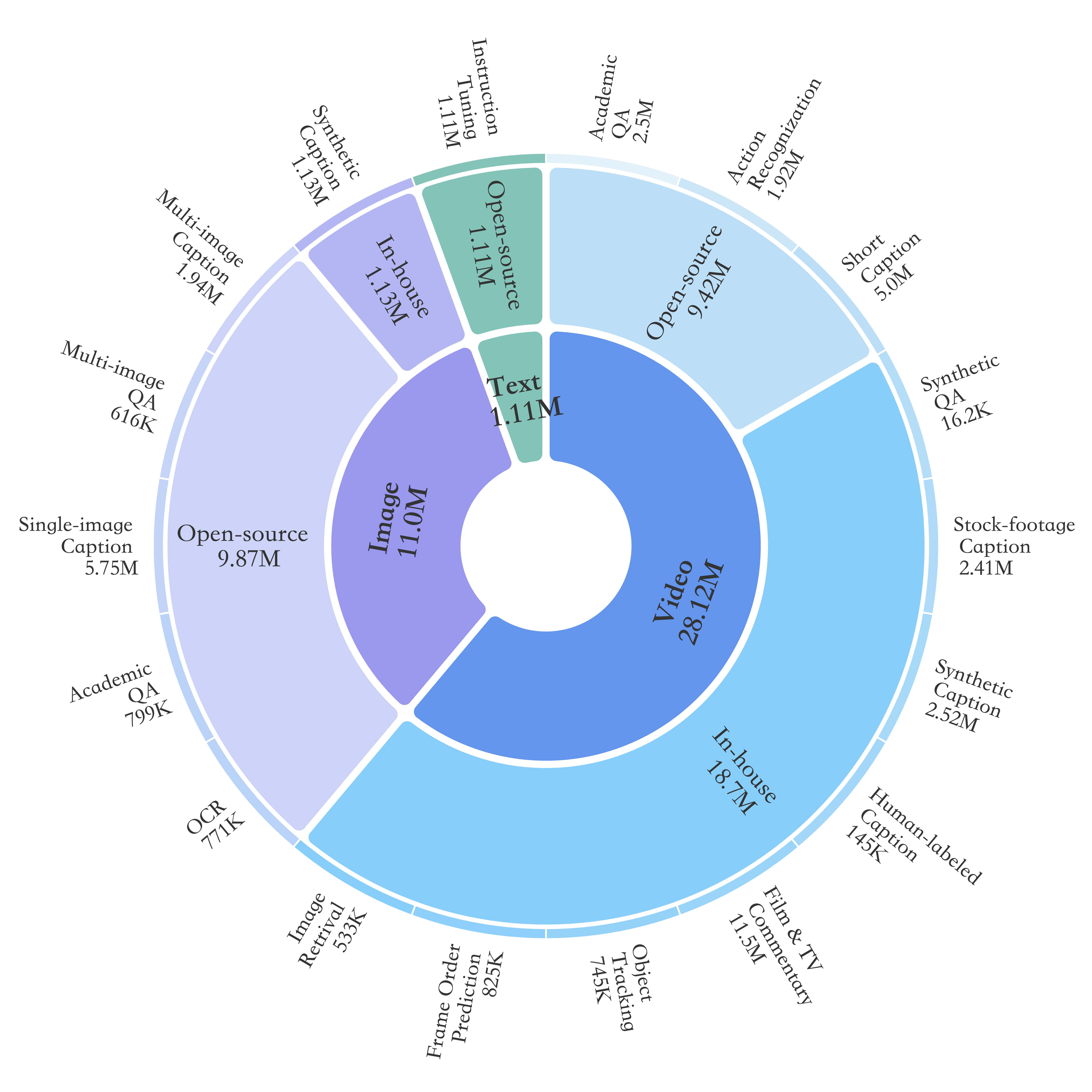}
    \caption{Summary of datasets used in the pre-training stage of \modelname.}
    \label{fig:pretrain_data}
\end{figure}

The pre-training stage encompasses a variety of tasks, including video captioning, video question answering, action recognition, action grounding, (multi-)image understanding, and text generation. The training data consists of 20 million public datasets and 20 million newly collected in-house datasets. Figure~\ref{fig:pretrain_data} illustrates the composition of the pre-training data, with a detailed breakdown presented in Appendix~\ref{sec:public-datasets}. Our findings indicate that the in-house data significantly enhances model's performance, complementing the public datasets. In the following, we describe the pipeline used for in-house data collection.

We collected a large group of videos from the Internet, spanning diverse genres such as animation, movies, TV series, short videos, stock footage, games and so on. The videos are categorized into three types:
\begin{itemize}
    \item {\bf Short videos with captions.} This category consists of 2.4 million videos directly sourced from the Internet, preserving their original video-caption pairs.
    \item {\bf Commentary videos for movies or TV shows.} The videos were segmented into single-shot clips using PySceneDetect\footnote{https://www.scenedetect.com/}. A filtering model removed static or low-quality clips. Adjacent clips were then merged to create continuous segments, ensuring final video durations ranged from 2 to 30 seconds. 
    We utilized an internal OCR tool to extract the commentary text from the video and use it as the caption. The areas containing the commentary text in the video were obscured. To ensure relevance, we trained a lightweight BERT-style\cite{kenton2019bert} model to filter out clips where the commentary lacked direct visual correspondence (e.g., character dialogues). This process produced 11.0 million video clips.
    \item {\bf Other videos.} These videos were processed similarly to the commentary videos, undergoing segmentation into shorter clips, filtering out low-quality clips, and merging adjacent clips. After this, we employed a multi-modal LLM to automatically generate video captions and question-answer pairs, resulting in a total of 2.7 million clips.
\end{itemize}

Commentary videos represent a significant portion of the pre-training data. Unlike traditional video-text datasets, such as HowTo100M~\cite{miech2019howto100m}, which rely on ASR transcripts, commentary data demonstrates stronger alignment between video and text. This commentary not only describes low-level visual elements, such as atomic actions, but also highlights high-level information like plot details. This type of data can substantially enhance the model's visual understanding at varying levels of granularity. 

In addition to video caption data, we incorporate large-scale synthetic datasets for tasks such as object tracking, frame order prediction, image retrieval, video question-answering, and image captioning during pre-training.

Overall, our pre-training dataset consists of 40 million samples. We trained \modelname on this dataset using 128 H100 GPUs, with all components of \modelname set to be trainable. For each video, we sampled between 16 and 128 frames, depending on its duration. In total, the pre-training stage of \modelname processed approximately 200 billion tokens.

\subsection{Supervised fine-tuning}

During the SFT phase, our primary objectives are to further improve the model's accuracy and comprehensiveness in video descriptions and ensure the outputs are human-like: well-structured, appropriately detailed, and capable of generating accurate long-form descriptions. To achieve this, we collected 150K video clips and conducted SFT in two stages.

\begin{figure}[t]
    \centering
    \includegraphics[width=\linewidth]{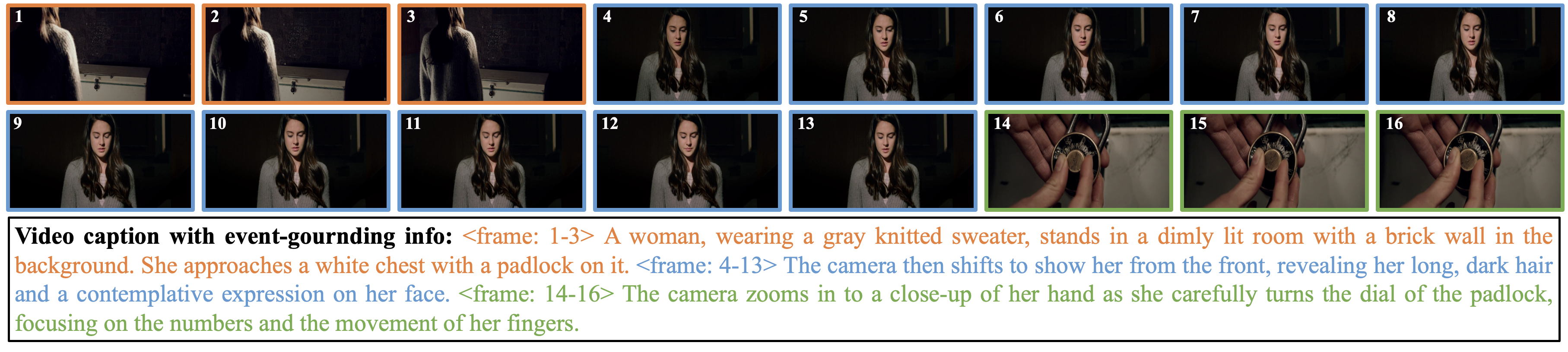}
    \caption{An example of a video description with fine-grained temporal grounding. ``$<$frame: $i$-$j>$" indicates that the following event is inferred from frames $i$ to $j$. Events are distinguished by color, with corresponding frames and descriptions marked in the same color to indicate their association.}
    \label{fig:sft_grounding}
\end{figure}

In the first stage, each video clip in the SFT dataset is annotated with a detailed description with fine-grained temporal grounding. As shown in Figure~\ref{fig:sft_grounding}, the annotations specify the frames corresponding to each event in the description. The annotation process is detailed in Appendix~\ref{sec:SFT-prompt}. This fine-grained frame-event alignment enhances the model’s ability to accurately identify and describe events by focusing on temporal and visual cues, complementing traditional video-caption alignment. Our experiments demonstrate that this approach mitigates the omission of key events in generated video descriptions.

In the second stage of SFT, we refined the model’s output to achieve a more human-like style. We observed that the data used in the initial stage of SFT often fragmented complete events into multiple steps due to event-grounding requirements. For instance, the action of pouring wine might be divided into steps like opening the bottle, lifting it, and pouring. To address this, we incorporated more natural and human-like video description data. Specifically, in this stage, we designed diverse description instructions to reflect real-world variations in language, granularity, and style requirements. We then annotated each video’s description to align with its corresponding instruction, as detailed in Appendix~\ref{sec:SFT-prompt}. This data allowed the model to better interpret varying instructions and generate more accurate and diverse video descriptions.

The training data for SFT-1 contains 150k video description pairs, while SFT-2 comprises 50k diverse instructions and 150k refined video-description pairs. Each pair includes a video description aligned with one of the instructions. We trained \modelname on this dataset using 32 H100 GPUs and set all components of \modelname to trainable. For each video, we sampled 16 frames for training. The global training batch size was set to 64, and \modelname was trained for 5000 iterations in each of the two phases. In addition, we used 2e-5 and 2e-6 as the learning rate of the model during the two-stage SFT respectively to obtain further performance improvement.

\subsection{Direct Preference Optimization}\label{sec:dpo}

In this subsection, we introduce a novel automated method for collecting preference data for video description. By performing DPO~\cite{rafailov2024direct} training on this data, we can further improve the model's ability to generate high-quality, detailed video descriptions.

\begin{figure}[ht]
    \centering
    \includegraphics[width=\linewidth]{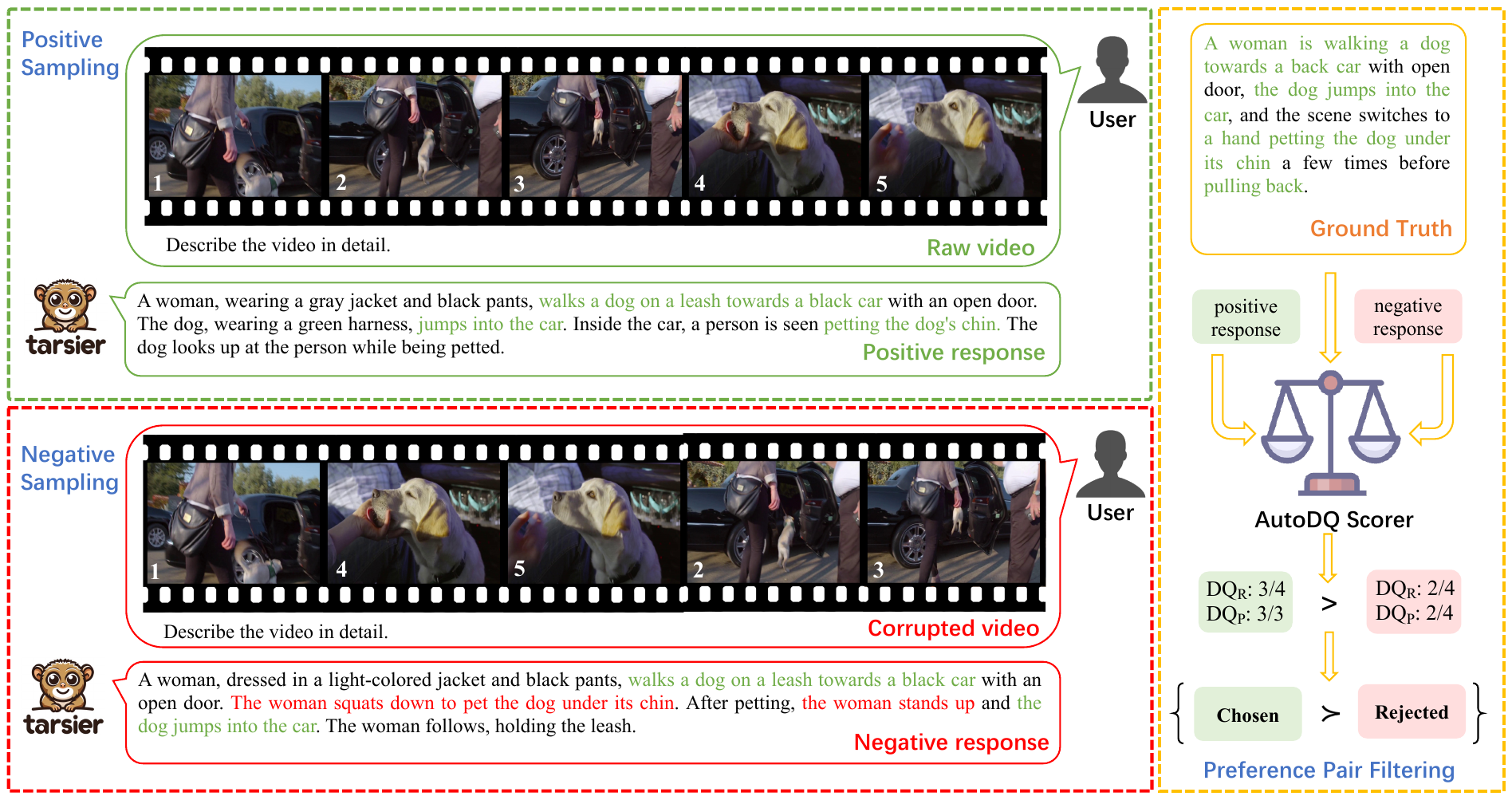}
    \caption{Preference data construction pipeline for DPO training.}
    \label{fig: preference_pair_pipeline}
\end{figure}

\paragraph{Negative sampling} Existing works often conduct multiple times sampling on the same input (video and text prompt) to acquire preference pair candidates\cite{xiong2024llavacriticlearningevaluatemultimodal, zhang2024directpreferenceoptimizationvideo, tang2024enhancingmultimodalllmdetailed}. In practice, however, we found that 1) Low-temperature sampling produces minimal variation in responses; 2) High-temperature sampling often leads to uncontrollable or abnormal generations. To address these issues, we propose a new automated preference data collection approach that enhances controllability and consistently yields high-quality preference data.

In reinforcement learning (RL) terms, the VLM serves as a policy model $\pi_\theta$, typically initialized from the SFT model. Given an input prompt $x$, consisting of $N$ frames sampled from a video, $\pi_\theta$ generates an video description $y$. Then, the video frames are modified to produce a corrupted prompt $\tilde{x}$ through one of the following perturbations:
\begin{itemize}
    \item Clip-switching: Evenly divide the video into 4 clips, then randomly choose 2 clips and swap their order.
    \item Clip-reversing: A random clip with $\frac{N}{2}\sim N$ frames is reversed.
    \item Clip-cropping: $N$ frames are resampled from a random clip with half of the video's original duration.
    \item Down-sampling: Half of the $N$ frames are randomly dropped.
\end{itemize}
The corrupted prompt $\tilde{x}$ is input into $\pi_{\theta}$, generating a new description $\tilde{y}$. The resulting preference data is represented as $\{x, y_w=y, y_l=\tilde{y}\}$. The first two perturbations are designed to induce negative descriptions with temporal errors, while the latter two are designed to induce incomplete descriptions. Consequently, through DPO training, the model can be enhanced to produce descriptions with improved accuracy and completeness.

Figure~\ref{fig: preference_pair_pipeline} provides an example to illustrate the preference data construction pipeline. From a raw video, we first generate a positive response using the current model. Next, a corrupted video, created through clip-switching, is fed into the model to obtain a negative sample, which contains two hallucinations (highlighted in red).

\paragraph{Preference data filtering} Given a prompt $x$, response $\tilde{y}$ is generally more negative compared to $y$. However, an effective filter mechanism for valid preference data remains essential, as $\tilde{y}$ is not always strictly worse than $y$\footnote{An obvious counter example is that a low-dynamic video will not be significantly affected by the down-sampling perturbation.}. As shown on the right side of Figure~\ref{fig: preference_pair_pipeline}, we utilize AutoDQ~\cite{wang2024tarsierrecipestrainingevaluating}, an automatic method for evaluating the quality of video description, using two metrics, $DQ_{R}$ and $DQ_P$\footnote{Given a reference description ($d_{ref}$) and a description to be assessed ($d_{pred}$), AutoDQ scorer outputs the recall score ($DQ_R$: the ratio of events in $d_{ref}$ that are entailed by $d_{pred}$) and the precision score ($DQ_P$: the ratio of events in $d_{ref}$ that are entailed by $d_{pred}$).}. A preference pair $\{x, y_w=y, y_l=\tilde{y}\}$ is considered valid if the following conditions are met:
\begin{equation}
\Delta DQ_{R} \geq 0  {\rm \quad and \quad} \Delta DQ_{P} \geq 0  {\rm \quad and \quad} \Delta DQ_{R} + \Delta DQ_{P} \geq \delta,
\end{equation}
where $\Delta DQ_{R}$ and $\Delta DQ_{P}$ denotes the difference of AutoDQ recall and precision scores between the $y_0$ and $y_1$. $\delta$ serves as an adjustable threshold to fine-tune the filtering criteria.

During the DPO training phase, we utilize videos from the same training dataset, $\mathcal{D}$, as in the SFT phase, to construct preference data. The policy model is then optimized by minimizing the DPO loss, expressed as:
\begin{equation}
\mathcal{L}_{DPO} = -\mathbb{E}_{(x,y_w,y_l)\sim\mathcal{D}}
\left[\log\sigma\left(\beta\log\frac{\pi_\theta(y_w|x)}{\pi_{\rm ref}(y_w|x)} - \beta\log\frac{\pi_\theta(y_l|x)}{\pi_{\rm ref}(y_l|x)}\right)\right], \label{eq:dpo_loss}
\end{equation}
where $\pi_{\rm ref}$ denotes the model obtained during the SFT phase.

We conducted DPO training on a dataset with 20k preference pairs produced by the above data collection approach, with all parameters set to be trainable. For each video, we sample 16 frames as same as the SFT phase. We trained \modelname for 1,000 steps in total with 64 H100 GPUs and each GPU loaded one pair at each training step, resulting in a global batch size of 64. See Appendix \ref{sec:dpo_ab_settings} for more details of DPO training.

\section{Experiments}
In this section, we first evaluate the model's performance on various video understanding benchmarks, comparing it to several baselines. We highlight \modelname's advantages not only in video description but also across other video understanding tasks. We then present an ablation study to examine key components of our approach.

\subsection{Quantitative Results}

\subsubsection{Video Captioning}
We evaluate \modelname on two video captioning benchmarks: DREAM-1K\cite{wang2024tarsierrecipestrainingevaluating} and E.T. Bench-Captioning\cite{liu2024etbench}. DREAM-1K is a detailed video description benchmark featuring dynamic and diverse videos, assessing the model's ability to describe fine-grained actions and events. E.T Bench-Captioning is composed of four dense video captioning tasks, requiring key event localization and summary generation for segments in long-form videos.

\begin{table}[h!]
\centering
\resizebox{\textwidth}{!}{
    \begin{tabular}{l|cccccc}
    \toprule
    \multirow{2}{*}{\textbf{Model}} & \multicolumn{5}{c}{\textbf{Video Categories}} & \multirow{2}{*}{\textbf{Overall}} \\
    & Live-action & Animation  & Stock & YouTube & Shorts & \\\midrule
    \multicolumn{7}{l}{\textit{Proprietary models}} \\
    GPT-4V \cite{gpt4v} & 34.8/39.2/31.3 & 27.4/31.9/24.0 & 40.7/\underline{46.7}/36.1 & 33.8/40.1/29.2 & 34.8/46.1/28.0 & 34.4/40.8/29.7 \\
    GPT-4o \cite{gpt4o} & 39.8/\underline{42.1}/37.8 & 35.8/39.1/33.1 & 44.0/46.6/41.7 & 35.9/\underline{41.5}/31.7 & 39.9/47.9/34.2 & 39.2/\underline{43.4}/35.7 \\
    Gemini-1.5-Flash \cite{geminiteam2024gemini15unlockingmultimodal} & 34.8/36.4/33.3 & 29.2/32.5/26.5 & 39.4/39.7/39.1 & 34.3/38.6/30.9 & 35.6/42.4/30.7 & 34.8/37.9/32.1 \\
    Gemini-1.5-Pro \cite{geminiteam2024gemini15unlockingmultimodal} & 36.4/36.4/36.4 & 30.7/31.8/29.7 & 42.2/40.7/43.8 & 34.0/36.7/31.6 & 37.0/42.4/32.7 & 36.2/37.6/34.8 \\
    \midrule
    \multicolumn{7}{l}{\textit{Open-source models ($>$10B)}} \\
    PLLaVA-34B \cite{xu2024pllava} & 29.3/34.9/25.2 & 20.9/32.0/15.6 & 35.1/42.5/29.9 & 28.9/40.8/22.3 & 25.6/41.9/18.4 & 28.2/38.4/22.3 \\
    VideoLLaMA2-72B \cite{cheng2024videollama2} & 27.3/29.3/25.6 & 19.7/21.7/18.1 & 33.9/37.0/31.3 & 27.7/33.0/23.8 & 26.5/33.1/22.1 & 27.1/30.8/24.2 \\
    LLaVA-OV-72B \cite{li2024llavanext} & 31.7/32.8/30.7 & 27.7/30.6/25.2 & 38.0/39.6/36.6 & 34.1/34.7/33.5 & 33.8/41.8/28.4 & 33.2/35.9/30.9 \\
    LLaVA-Video-72B \cite{zhang2024video} & 33.5/36.3/31.1 & 28.6/31.7/26.1 & 39.3/41.1/37.6 & 32.8/34.7/31.1 & 35.7/42.8/30.6 & 34.0/37.3/31.3\\
    Qwen2-VL-72B \cite{qwen2vl} & 32.1/33.7/30.6 & 27.6/32.6/23.9 & 41.1/41.2/41.1 & 32.0/38.1/27.7 & 32.1/41.0/26.4 & 33.2/37.3/29.9\\
    InternVL2.5-78B \cite{chen2024expanding} & 25.3/31.5/21.1 & 21.8/28.8/17.6 & 33.5/38.1/29.9 & 31.0/38.5/25.9 & 31.1/41.7/24.8 & 28.6/35.7/23.9\\
    Tarsier-34B \cite{wang2024tarsierrecipestrainingevaluating} & 38.5/39.6/37.5 & 32.2/35.8/29.2 & 41.7/46.4/37.8 & 34.5/41.1/29.7 & 34.0/44.1/27.7 & 36.3/41.4/32.4  \\
    \midrule
    \multicolumn{7}{l}{\textit{Open-source models ($<$10B)}} \\
    Video-LLaVA-7B \cite{lin2023video} & 19.4/24.3/16.2 & 15.3/21.2/11.9 & 27.0/33.5/22.7 & 21.2/31.9/15.8 & 18.5/29.4/13.5 & 20.4/28.1/16.0 \\
    VideoLLaMA2-7B \cite{cheng2024videollama2} & 25.1/28.7/22.2 & 20.4/25.5/17.0 & 32.6/35.5/30.2 & 27.5/33.5/23.4 & 24.5/34.1/19.2 & 26.2/31.5/22.4 \\
    LLaVA-OV-7B \cite{li2024llavanext} & 31.2/33.2/29.3 & 26.8/29.0/25.0 & 38.1/39.1/37.1 & 30.6/32.1/29.2 & 31.4/38.3/26.6 & 31.7/34.3/29.4 \\
    LLaVA-Video-7B \cite{zhang2024video} & 31.4/35.2/28.4 & 27.6/32.9/23.8 & 36.7/39.7/34.1 & 33.0/\textbf{39.5}/28.3 & 33.4/42.5/27.5 & 32.5/37.9/28.4 \\
    Qwen2-VL-7B \cite{qwen2vl} & 27.7/32.5/24.2 & 22.2/28.0/18.4 & 37.0/36.1/38.0 & 30.7/35.5/27.0 & 29.1/37.6/23.8 & 29.6/33.9/26.3 \\ 
    InternVL2.5-8B \cite{chen2024expanding} & 26.6/32.0/22.8 & 21.3/28.9/16.9 & 32.7/37.2/29.1 & 27.9/35.4/23.0 & 28.9/39.9/22.7 & 27.6/34.7/22.9 \\
    Tarsier-7B \cite{wang2024tarsierrecipestrainingevaluating} & 36.6/38.5/34.8 & 29.3/34.6/25.5 & 39.6/44.7/35.5 & 33.0/39.2/28.4 & 33.6/44.6/26.9 & 34.6/40.3/30.2 \\
    \midrule
    Tarsier2-7B & \underline{\textbf{44.4}}/\textbf{41.9}/\underline{\textbf{47.3}} & \underline{\textbf{39.3}}/\underline{\textbf{39.5}}/\underline{\textbf{39.1}} & \underline{\textbf{45.7}}/\textbf{45.4}/\underline{\textbf{46.0}} & \underline{\textbf{36.0}}/38.4/\underline{\textbf{33.9}} & \underline{\textbf{43.7}}/\underline{\textbf{48.9}}/\underline{\textbf{39.4}} & \underline{\textbf{42.0}}/\textbf{42.8}/\underline{\textbf{41.1}} \\
    \bottomrule
    \end{tabular}
}
\caption{Evaluation results on DREAM-1K. We report F1/Precision/Recall scores for each category and for the overall dataset. For open-source models, all results are tested with their official checkpoint and inference code under recommended setting. SOTA results of comparable scale ($<$10B) are bolded and overall best results are underlined.}
\label{tab:dream-1k}
\end{table}

\begin{figure}[t]
    \centering
    \includegraphics[width=0.6\linewidth]{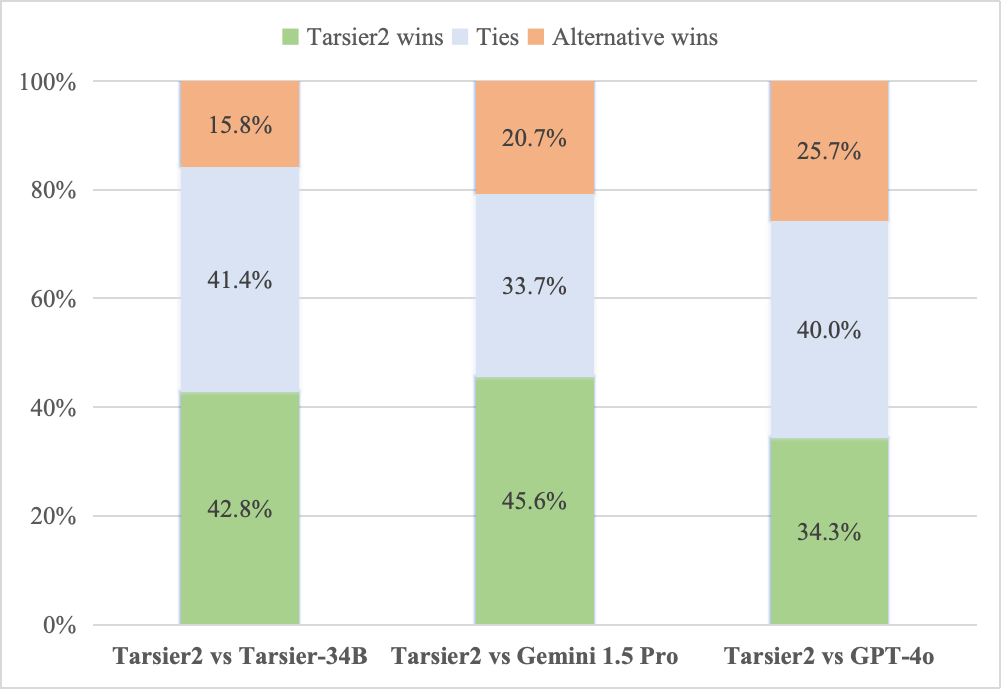}
    \caption{Human side-by-side evaluation results of Tarsier2 versus other models.}
    \label{fig:human_sbs}
\end{figure}

As shown in Table \ref{tab:dream-1k}, \modelname-7B outperforms all open-source models in both precision and recall across all categories in DREAM-1K, demonstrating its ability to generate more comprehensive and less hallucinatory video descriptions. Notably, \modelname-7B achieved an overall F1 score of 42.0\%, surpassing the strongest proprietary model, GPT-4o (39.2\%). It is also the first model to exceed a 40\% overall recall score, highlighting its sensitivity to dynamic actions and events.

Figure \ref{fig:human_sbs} further presents the human side-by-side evaluation results of Tarsier2 versus the previous SOTA Tarsier-34B and two strong proprietary models, GPT-4o and Gemini 1.5 Pro. We randomly sampled 250 videos (50 videos for each category) from DREAM-1K, and asked experienced annotators to compare the descriptions generated by two different models, collecting their preferences. Each pair of descriptions was randomly shuffled to ensure that the annotators were blind to the description sources. Compared to Tarsier-34B, Tarsier2 has a slightly negative advantage rate (15.8\%), but wins in a significant percentage of cases (42.8\%). Compared to Gemini, Tarsier2 still maintains a significant advantage (45.6\% vs 20.7\%). Despite being tied with the strongest proprietary model, GPT-4o, on 40\% cases, Tarsier2 still gains a slight advantage (8.6\%), demonstrating the outstanding performance of Tarsier2 in detailed video description. For a comparison of generated descriptions from different models on DREAM-1K, see Appendix \ref{sec:dream_cases}.

Table \ref{tab:dense_caption} shows the evaluation results of dense video captioning on E.T. Bench-Captioning. \modelname-7B outperforms all open-source models with comparable settings (similar model scale, fine-tuned on E.T. Instruct 164K~\cite{liu2024etbench}) across all metrics,  except for the SLC$_{F1}$ score, which is slightly lower than Qwen2-VL-7B (24.6\% vs 25.7\%). These results highlight \modelname's strengths in generating fine-grained descriptions for short videos and providing coarse-grained summaries for long videos.

\begin{table}[h!]
    \centering
    \resizebox{0.8\textwidth}{!}{%
    \begin{tabular}{l|cccccc}
    \toprule
        \multirow{2}{*}{\textbf{Model}} & \multicolumn{6}{c}{\textbf{E.T. Bench-Captioning} \cite{liu2024etbench}} \\
        & DVC$_{F1}$ & DVC$_{Sim}$ & SLC$_{F1}$ & SLC$_{Sim}$ & \textbf{Avg}$_{F1}$ & \textbf{Avg}$_{Sim}$\\
    \midrule
    \multicolumn{5}{l}{\textit{Proprietary models}} \\
        GPT-4V \cite{gpt4v} & \color{lightgray}16.1 & \color{lightgray}19.4 & \color{lightgray}21.9 & \color{lightgray}13.5 & \color{lightgray}19.0 & \color{lightgray} 16.4\\
        GPT-4o \cite{gpt4o} & \color{lightgray}\underline{46.9} & \color{lightgray}22.3 & \color{lightgray}23.1 & \color{lightgray}14.9 & \color{lightgray}35.0 & \color{lightgray}18.6\\
        Gemini-1.5-Flash \cite{geminiteam2024gemini15unlockingmultimodal} & \color{lightgray}31.6 & \color{lightgray}14.9 & \color{lightgray}16.5 & \color{lightgray}13.3 & \color{lightgray}24.1 & \color{lightgray}14.1\\
        Gemini-1.5-Pro \cite{geminiteam2024gemini15unlockingmultimodal} & \color{lightgray}24.0 & \color{lightgray}17.5 & \color{lightgray}5.8 & \color{lightgray}9.8 & \color{lightgray}14.9 & \color{lightgray}13.7\\
    \midrule
    \multicolumn{5}{l}{\textit{Open-source models ($>$10B)}} \\
        PLLaVA-34B \cite{xu2024pllava} &13.3 & 10.6 & 9.7 & 11.8 & 11.5 & 11.2 \\
        LLaVA-OV-72B \cite{li2024llavanext} & 41.9 & 16.3 & 25.6 & 13.9 & 33.8 & 15.1 \\
        LLaVA-Video-72B \cite{zhang2024video} & 37.0 & 15.7 & 20.4 & 13.5 & 28.7 & 14.6 \\   
        Qwen2-VL-72B \cite{qwen2vl} & 15.3 & 13.9 & 11.0& 12.8 & 13.2 & 13.4 \\
    \midrule
    \multicolumn{5}{l}{\textit{Open-source models ($\leq$10B)}} \\
       VideoLLaMA2-7B \cite{cheng2024videollama2} & 0.6 & 14.5 & 0.0 & 15.2 & 0.3 & 14.8 \\
       Video-LLaVA-7B \cite{lin2023video} & 28.0 & 15.0 & 0.9 & 8.3 & 14.4 & 11.7\\
       LLaVA-OV-7B \cite{li2024llavanext} & 22.0 & 15.1 & 9.5 & 10.6 & 15.8 & 12.8 \\
       LLaVA-Video-7B \cite{zhang2024video} & 20.6 & 14.7 & 6.5 & 13.4 & 13.6 & 14.1 \\
       E.T. Chat \cite{liu2024etbench} $^\dag$ & 38.4 & 19.7 & 24.4 & 14.6 & 31.4 & 17.1 \\
       Qwen2-VL-7B \cite{qwen2vl} $^\dag$ & 44.3 & 25.3 & \underline{\textbf{25.7}} & 15.6 & 35.0 & 20.4 \\      
       Tarsier-7B \cite{wang2024tarsierrecipestrainingevaluating} $^\dag$ & 42.8 & 19.1 & 23.7 & 15.2 & 33.2 & 17.1 \\
    \midrule
        Tarsier2-7B $^\dag$& \textbf{46.5} & \underline{\textbf{28.8}} & 24.6 & \underline{\textbf{16.4}} & \underline{\textbf{35.5}} & \underline{\textbf{22.6}} \\
    \bottomrule
    \end{tabular}%
    }
    \caption{Evaluation results on E.T. Bench-Captioning. Results marked in gray are tested on a subset. $\dag$ denotes the model is fine-tuned on E.T. Instruct 164K. All results are transcribed from the official benchmark, except for LLaVA-OV, LLaVA-Video and Qwen2-VL, which are our evaluation using the official checkpoint and inference code.}
    \label{tab:dense_caption}
\end{table}

\subsubsection{Short-Video Question Answering}

\begin{table}[h!]
    \centering
    \resizebox{0.98\textwidth}{!}{%
    \begin{tabular}{l|cccccc}
    \toprule
        \multirow{2}{*}{\textbf{Model}} & \textbf{MVBench}\cite{li2024mvbench} & \textbf{PerceptionTest}\cite{patraucean2024perception} & \textbf{TVBench}\cite{cores2024tvbench} & \textbf{TOMATO}\cite{shangguan2024tomato} & \textbf{Vinoground}\cite{zhang2024vinoground} & \textbf{TempCompass}\cite{liu2024tempcompass} \\
        & test & val & test & test & Text/Video/Group & mc/yn/cm/cg\\
    \midrule
    \multicolumn{6}{l}{\textit{Proprietary models}} \\
        GPT-4o \cite{gpt4o} & 57.5 & -&39.6 & 37.7 & 54.0/\underline{38.2}/24.6 & 71.0/73.7/80.8/70.8 \\
        Gemini-1.5-Pro \cite{geminiteam2024gemini15unlockingmultimodal} & -& -& 46.5 & 36.1 & 35.8/22.6/10.2 & 63.9/70.3/77.5/57.9\\
    \midrule
    \multicolumn{6}{l}{\textit{Open-source models ($>$10B)}} \\
        LLaVA-OV-72B \cite{li2024llavanext} & 59.4& 66.9& 45.9 & 28.6 & 48.4/35.2/21.8 & 67.6/72.6/78.2/52.6 \\
        LLaVA-Video-72B \cite{zhang2024video} & 64.1& 
\underline{74.3}*& 50.0 & 28.2 & 52.0/35.6/20.8 & 69.9/73.0/80.9/54.4 \\
        Qwen2-VL-72B \cite{qwen2vl} & 
\underline{73.6} & 66.5 & 52.7 & 37.9 & 50.4/32.6/17.4 & \underline{76.0}/\underline{75.9}/\underline{84.6}/58.6 \\
        Tarsier-34B \cite{wang2024tarsierrecipestrainingevaluating} & 67.6&60.4&53.8&34.3&37.8/32.0/15.0 & 69.8/74.0/73.0/60.9 \\
    \midrule
    \multicolumn{6}{l}{\textit{Open-source models ($\leq$10B)}} \\
        LLaVA-OV-7B \cite{li2024llavanext} & 56.7& 57.1& 45.6 & 25.5& 41.6/29.4/14.6 & 64.8/69.7/73.8/49.9\\
        LLaVA-Video-7B \cite{zhang2024video} & 58.6 & 67.9*& 45.6 & 24.9 & 36.8/29.0/12.8 & 56.3/68.7/76.8/53.0\\
        Qwen2-VL-7B \cite{qwen2vl} & 67.0& - & 43.8 & 31.5& 40.0/23.4/12.4 & 68.5/72.8/77.3/54.2\\
        Tarsier-7B \cite{wang2024tarsierrecipestrainingevaluating} & 62.6&53.9&45.8&28.6&29.8/22.2/8.6 & 58.7/58.0/54.2/55.3\\
        Previous SOTA & \textbf{72.0} \cite{chen2024expanding} & 70.0* \cite{liu2024oryx} & 51.6 \cite{zhang2024internlmxcomposer} & 31.5 \cite{qwen2vl} & 41.6/29.4/14.6 \cite{li2024llava} &
        68.5/72.8/77.3/54.2 \cite{qwen2vl}\\
    \midrule
        Tarsier2-7B & 71.5& \textbf{71.6}* & \underline{\textbf{54.7}} & \underline{\textbf{42.0}} & \underline{\textbf{65.8}}/\textbf{38.0}/\underline{\textbf{28.8}} & \textbf{75.3}/\textbf{75.1}/\textbf{80.6}/\underline{\textbf{66.6}}\\
    \bottomrule
    \end{tabular}%
    }
    \caption{Evaluation results on short video question answering benchmarks. * indicates that the training set has been observed in the training data mixture.}
    \label{tab:my_label}
\end{table}

We evaluate \modelname-7B on several short-video question answering benchmarks to assess its ability to comprehend and reason about visual content. As shown in Table \ref{tab:my_label}, \modelname-7B outperforms both proprietary and open-source models across various benchmarks, achieving state-of-the-art results. \modelname-7B exhibits exceptional performance in MVBench \cite{li2024mvbench} and PerceptionTest \cite{patraucean2024perception}, with scores of 71.5\% and 71.6\%, respectively.

Furthermore, \modelname-7B demonstrates significant performance improvements on benchmarks featuring temporal reasoning, such as TVBench \cite{cores2024tvbench}, TOMATO \cite{shangguan2024tomato}, and Vinoground \cite{zhang2024vinoground}. \modelname-7B achieves strong results with 54.7\% on TVBench, 42.0\% on TOMATO, and 65.8\%/38.0\%/28.8\% on Vinoground's Text/Video/Group tasks, respectively. These results surpass both open-source and proprietary models, including GPT-4o and Gemini-1.5-Pro. 

At last, \modelname-7B also excels on the TempCompass benchmark \cite{liu2024tempcompass}, which evaluates temporal perception in ten aspects and four task formats. \modelname-7B achieves impressive scores 
of 75.3\%/75.1\%/80.6\%/66.6\% on TempCompass' mc/yn/cm/cg tasks, respectively, outperforming both open-source models and larger proprietary models in most cases. This performance further underscores \modelname's advanced ability to process and interpret temporal information in video content.

\subsubsection{Long-Video Question Answering}

\begin{table}[h!]
    \centering
    \resizebox{0.98\textwidth}{!}{%
    \begin{tabular}{l|ccccc}
    \toprule
        \multirow{2}{*}{\textbf{Model}} & \textbf{Video-MME}\cite{fu2024video} & \textbf{LongVideoBench}\cite{wu2024longvideobench} & \textbf{TemporalBench}\cite{cai2024temporalbench} & \textbf{MLVU}\cite{zhou2024mlvu} &\textbf{MMBench-Video}\cite{fang2024mmbench}\\
         & w/o subs & val & Binary Accuracy & M-Avg & val \\
    \midrule
    \multicolumn{6}{l}{\textit{Proprietary models}} \\
    GPT-4o \cite{gpt4o} &  71.9 & \underline{66.7} & \underline{73.2} & 64.6 & 1.87 \\
    Gemini-1.5-Pro \cite{geminiteam2024gemini15unlockingmultimodal} & \underline{75.0} & 64.0 & 66.4 & - & 1.30 \\
    
    \midrule
    \multicolumn{6}{l}{\textit{Open-source models ($>$10B)}} \\
    VILA-1.5-40B \cite{lin2024vila} & 60.1 & - & - & 56.7 & 1.61 \\
    LLaVA-Video-72B \cite{zhang2024video} & 70.5 & 61.9 & 72.4 & 74.4 & 1.71 \\
    Qwen2-VL-72B \cite{qwen2vl} & 71.2 & - & 70.2 & - & 1.70 \\
    InternVL2.5-78B \cite{chen2024expanding} & 72.1 & 63.6 & - & \underline{75.7} & 
\underline{1.97} \\
    Tarsier-34B \cite{wang2024tarsierrecipestrainingevaluating} & 52.3 & 54.2 & 66.7 & 58.2 & 1.46 \\
    \midrule
    \multicolumn{6}{l}{\textit{Open-source models ($\leq$10B)}} \\
    LLaVA-Video-7B \cite{zhang2024video} & 63.3 & 58.2 & 63.6 & 70.8 & 1.60 \\
    Qwen2-VL-7B \cite{qwen2vl} & 63.3 & 55.6 & 62.0 & - & 1.44 \\
    InternVL2.5-8B \cite{chen2024expanding} & 64.2 & 60.0 & - & 68.9 & 1.68 \\ 
    Tarsier-7B \cite{wang2024tarsierrecipestrainingevaluating} & 42.2 & 39.8 & 56.9 & 49.3 & - \\
    Previous SOTA & 64.2 \cite{liu2024nvila} & 60.0 \cite{chen2024expanding} & 63.6 \cite{zhang2024video} & 70.9 \cite{zohar2024apolloexplorationvideounderstanding} & 1.70 \cite{yao2024minicpm} \\
    \midrule
    \modelname-7B & \textbf{64.5} (128f) & 58.6 (128f) & \textbf{65.3} (128f) & 67.9 (256f) & \textbf{1.82} (128f) \\
    \bottomrule
    \end{tabular}%
    }
    \caption{Evaluation results on long-video question answering benchmarks. We list the number of frames used for each benchmark during evaluating \modelname.}
    \label{tab:results-long-video-qa}
\end{table}

We evaluate \modelname on long-video question answering benchmarks by uniformly sampling 128 or 256 frames, depending on the video length. Comparison results with other proprietary and open-source models are presented in Table~\ref{tab:results-long-video-qa}. Despite our training set not including many long video data, \modelname, compared with others under 10 billion parameters, still achieves SOTA on three benchmarks and competitive performance on several other benchmarks.


\subsubsection{Hallucination}

\begin{table}[h!]
\centering
\resizebox{\textwidth}{!}{%
\begin{tabular}{l|c|cc}
\toprule
\multirow{2}{*}{\textbf{Model}} & \textbf{VideoHallucer} \cite{wang2024videohallucer} & \multicolumn{2}{c}{\textbf{EventHallusion }\cite{zhang2024eventhallusion}} \\
\cline{2-4}
& Yes/No QA & Yes/No QA & Desc GPT \\
& Basic/Hallucinated/\textbf{Overall} & Entire/Interleave/Misleading/\textbf{Overall} & Entire/Interleave/Misleading/\textbf{Overall} \\
\midrule
\multicolumn{4}{l}{\textit{Proprietary models}}\\
GPT-4o \cite{gpt4o} & 75.1/74.2/53.3 & 65.8/90.7/92.2/84.1 & 34.9/54.9/83.2/56.2 \\
Gemini-1.5-Pro \cite{geminiteam2024gemini15unlockingmultimodal} & 83.6/42.3/37.8 & 70.2/77.7/96.1/80.2 & 38.5/40.9/80.0/49.6 \\
\midrule
\multicolumn{4}{l}{\textit{Open-Source models ($>$10B)}} \\
Qwen2-VL-72B \cite{qwen2vl} & 87.1/79.4/\underline{70.2} & 33.3/77.7/56.4/60.0 & 16.5/25.4/70.2/33.6 \\
LLaVA-OV-72B \cite{li2024llavanext} & 88.3/62.6/55.2 & 47.4/26.9/90.1/48.3 & 24.8/34.7/71.3/40.7 \\
LLaVA-Video-72B \cite{zhang2024video} & 88.2/73.5/64.6  & 57.9/11.9/96.0/45.6 & 32.1/35.8/75.5/44.2 \\
InternVL2.5-78B \cite{chen2024expanding} & 82.5/82.5/67.8 & 57.9/67.9/88.2/70.2 & 45.0/43.0/76.8/51.6 \\
Tarsier-34B \cite{wang2024tarsierrecipestrainingevaluating} & 84.8/80.0/67.7 & 49.1/92.7/69.6/74.8 & 38.5/40.4/83.2/50.1 \\
\midrule
\multicolumn{4}{l}{\textit{Open-Source models ($\leq$10B)}} \\
LLaVA-OV-7B \cite{li2024llavanext} & 81.1/69.6/53.8 & 46.5/67.4/86.1/66.2 & 22.0/26.4/73.4/36.4 \\
LLaVA-Video-7B \cite{zhang2024video} & 82.4/70.6/56.0 & 61.4/48.7/96.0/64.0 & 27.5/32.6/75.5/41.4 \\
Qwen2-VL-7B \cite{qwen2vl} & 85.0/70.8/59.3 & 35.1/94.3/57.4/68.6 & 14.7/16.1/67.0/27.8 \\
InternVL2.5-8B \cite{chen2024expanding} &72.7/78.3/53.6&46.5/69.2/90.2/68.2&23.9/20.7/60.0/31.0\\
Tarsier-7B \cite{wang2024tarsierrecipestrainingevaluating} & 76.4/60.8/41.4 & 43.9/82.4/79.4/70.9 & 35.8/29.5/72.6/41.6 \\
\midrule
\modelname-7B & 86.5/78.3/\textbf{67.0}&60.5/93.3/95.1/\underline{\textbf{84.6}}&54.6/53.1/93.7/\underline{\textbf{63.3}} \\
\bottomrule
\end{tabular}}
\caption{Evaluation results on hallucination benchmarks.}
\label{tab:video_hallucination}
\end{table}

We evaluate \modelname on two video hallucination benchmarks: VideoHallucer \cite{wang2024videohallucer} and EventHallusion \cite{zhang2024eventhallusion}. The results are summarized in Table \ref{tab:video_hallucination}. For VideoHallucer, \modelname-7B achieves an overall score of 67.0\%, outperforming all comparable baselines of similar model scale and even proprietary models like GPT-4o and Gemini-1.5-pro. In EventHallusion, for video question-answering task, \modelname-7B achieves 84.6\%, surpassing GPT-4o’s score of 84.1\%, while outperforming all other baselines. For the detailed description matching task, which directly assesses video description hallucinations by prompting GPT-4 to answer questions based on each model's generated video description, \modelname-7B demonstrates superior performance, even surpassing GPT-4o by 7.1\% in terms of Overall score.


\subsubsection{Video Grounding}

\begin{table}[h]
    \centering
    \resizebox{0.8\textwidth}{!}{%
    \begin{tabular}{l|cccccc}
    \toprule
        \multirow{2}{*}{\textbf{Model}} & \multicolumn{6}{c}{\textbf{E.T. Bench-Grounding} \cite{liu2024etbench}}\\
        & TVG$_{F1}$ & EPM$_{F1}$ & TAL$_{F1}$ & EVS$_{F1}$ & VHD$_{F1}$ & \textbf{Mean$_{F1}$}\\
    \midrule
    \multicolumn{7}{l}{\textit{Proprietary models}} \\
        GPT-4V \cite{gpt4v} & \color{lightgray}27.0 & \color{lightgray}1.8 & \color{lightgray}18.0 & \color{lightgray}\underline{28.6} & \color{lightgray}55.1 & \color{lightgray}26.1\\
        GPT-4o \cite{gpt4o} & \color{lightgray}40.4 & \color{lightgray}4.5 & \color{lightgray}20.0 & \color{lightgray}17.6 & \color{lightgray}56.9 & \color{lightgray}27.9\\
        Gemini-1.5-Flash \cite{geminiteam2024gemini15unlockingmultimodal} & \color{lightgray}\underline{43.9} & \color{lightgray}5.4 & \color{lightgray}27.0 & \color{lightgray}5.4 & \color{lightgray}60.8 & \color{lightgray}28.5\\
        Gemini-1.5-Pro \cite{geminiteam2024gemini15unlockingmultimodal} & \color{lightgray}43.1  & \color{lightgray}6.2 & \color{lightgray}33.8 & \color{lightgray}7.9 & \color{lightgray}47.0 & \color{lightgray}27.6\\
    \midrule
    \multicolumn{7}{l}{\textit{Open-source models ($<$10B)}} \\
       LITA \cite{huang2024lita} & 22.2 & 4.6 & 18.0 & 29.7 & 23.9 & 19.7 \\
       VTG-LLM \cite{guo2024vtg} & 15.9 & 3.7 & 14.4 & 26.8 & 48.2 & 21.8 \\
       TimeChat \cite{Ren2023TimeChat} $^\dag$ & - & - & - & - & - & 24.3 \\
       E.T. Chat \cite{liu2024etbench} $^\dag$ & 38.6 & 10.2 & 30.8 & \textbf{25.4} & 62.5 & 33.5 \\
       Tarsier-7B \cite{wang2024tarsierrecipestrainingevaluating} $^\dag$ & 39.6 & 9.0 & 25.0 & \textbf{25.4} & 47.6 & 30.9 \\
       Qwen2-VL-7B \cite{qwen2vl} $^\dag$ & \textbf{39.7} & 7.0 & 26.9 & 17.1 & \underline{\textbf{66.9}} & 33.5 \\
    \midrule
        Tarsier2-7B $^\dag$ & 38.4 & \underline{\textbf{11.0}} & \underline{\textbf{31.8}} & 19.4 & 66.8 & \underline{\textbf{35.5}} \\
    \bottomrule
    \end{tabular}
    }
    \caption{Evaluation results on E.T. Bench-Grounding. Results marked in gray are tested on a subset. $\dag$ denotes the model is fine-tuned on E.T. Instruct 164K.}
    \label{tab:grounding}
\end{table}

We evaluate the video grounding capability of models on E.T. Bench-Grounding, which combines various grounding tasks from multiple datasets, including QVHighlights \cite{lei2021detecting}, Charades-STA \cite{gao2017tall}, THUMOS'14 \cite{idrees2017thumos}, and Ego4D-NLQ \cite{grauman2022ego4d}, among others. The results, shown in Table~\ref{tab:grounding}, indicate that \modelname-7B achieves the highest mean F1 score of 35.5\%, outperforming all baselines and highlighting its superior temporal perception capabilities.


\subsubsection{Embodied Question Answering}

\begin{table}[h]
    \centering
    \setlength{\tabcolsep}{2pt} 
    \scriptsize
    \begin{minipage}{0.27\textwidth}
        \centering
        \begin{tabular}{l|c}
            \toprule
             \multirow{2}{*}{\textbf{Model}} & \textbf{EgoTaskQA} \\
             & Exact Match \\
            \midrule
            Human & 80.0 \\
            HCRN \cite{le2020hierarchical} & 42.2 \\
            GF \cite{bai2024glance} & 44.3 \\
            EgoVLPv2 \cite{pramanick2023egovlpv2} & 46.3 \\
            \midrule
            \modelname & \underline{\textbf{77.5}} \\
            \bottomrule
        \end{tabular}
    \end{minipage}
     \begin{minipage}{0.415\textwidth}
        \centering
        \begin{tabular}{l|c}
            \toprule
            \multirow{2}{*}{\textbf{Model}} & \textbf{RoboVQA} \\
            & BLEU-1/2/3/4\\
            \midrule
            LLaMA-AdapterV2 \cite{gao2023llama} & 27.8/16.0/10.9/8.1 \\
            LLaVA-OV-7B \cite{li2024llavanext} & 38.1/33.6/31.8/31.0 \\
            RoboMamba \cite{liu2024robomamba} & 54.9/44.2/39.5/36.3 \\
            MLCD \cite{an2025multi} & 73.2/66.4/60.6/56.6 \\
            \midrule
            \modelname & \underline{\textbf{77.1}}/\underline{\textbf{67.4}}/\underline{\textbf{61.5}}/\underline{\textbf{56.8}} \\
            \bottomrule
        \end{tabular}
    \end{minipage}
    \begin{minipage}{0.3\textwidth}
        \centering
        \begin{tabular}{l|c}
            \toprule
            \multirow{2}{*}{\textbf{Model}} & \textbf{OpenEQA} \\
            & GPT-4\\
            \midrule
            Human & 86.8 \\
            GPT-4V \cite{gpt4v} & 55.3 \\
            Gemini-1.5-Pro \cite{geminiteam2024gemini15unlockingmultimodal} & 44.9 \\
            MLCD \cite{an2025multi} & 48.8 \\
            \midrule
            \modelname & \underline{\textbf{58.7}} \\
            \bottomrule
        \end{tabular}
    \end{minipage}
    \caption{Evaluation results on embodied question-answering tasks, including EgoTaskQA, RoboVQA and OpenEQA.}
    \label{table:evaluate-EgoTaskQA}
\end{table}

We evaluate \modelname on embodied question answering to assess its performance in real-world robotic scenarios, using three benchmarks: EgoTaskQA \cite{jia2022egotaskqa}, RoboVQA \cite{robovqa2023arxiv}, and OpenEQA \cite{majumdar2024openeqa}.  To align with the baselines, \modelname is fine-tuned on the training sets for EgoTaskQA and RoboVQA, while for OpenEQA, it is evaluated in a zero-shot setting. The results, presented in Table \ref{table:evaluate-EgoTaskQA}, include exact match accuracy for EgoTaskQA, BLEU score for RoboVQA, and the correctness score evaluated by GPT-4-1106-preview \cite{gpt4} for OpenEQA. \modelname achieves top-tier performance across all three benchmarks, outperforming both generalist and specialist models. Notably, on EgoTaskQA, its performance approaches human-level accuracy, highlighting the model's significant potential in embodied intelligence.

\subsection{Ablation Study}
We conduct a comprehensive ablation study to evaluate key components at different stages of the training process. The study is based on three tasks: 1) {\bf Caption}: This includes the DREAM-1K dataset, the caption generation task from TempCompass (TempCompass-cg), and the caption matching task from Vinoground (Vinoground-Text) to assess captioning performance. 2) {\bf Video QA}: This encompasses short-video QA, measured by the average accuracy on MVBench, TVBench, and TOMATO, and long-video QA, measured by the average accuracy on Video-MME, LongVideoBench, and TemporalBench. It evaluates the model's video understanding capabilities. 3) {\bf Hallucination}: We use the average score of two sub-tasks from EventHallusion to assess hallucination in the model. The following subsections present the results for each task, with detailed results for individual datasets provided in the Appendix~\ref{sec:ab_exp_detail}.

\subsubsection{Pre-training}

\begin{table}[h]
    \centering
    \scriptsize
    \setlength{\tabcolsep}{3pt} 
    \resizebox{\textwidth}{!}{%
    \begin{tabular}{l|ccc|cc|c}
    \toprule
    \multirow{2}{*}{\textbf{Model}} & \multicolumn{3}{c|}{\textbf{Caption}} & \multicolumn{2}{c|}{\textbf{Video QA}} & \multirow{2}{*}{\textbf{Hallucination}} \\
    & DREAM-1K & TempCompass-cg & Vinoground-Text & Short & Long & \\
    \midrule
    Tarsier1-7B & 34.6 & 55.3 & 29.8 & 45.6 & 46.3 & 56.3 \\
    \midrule
    \makecell[l]{Tarsier1-7B-Qwen\\ \quad\textit{upgrading model}} & 38.4 ($\uparrow$3.8) & 59.3 ($\uparrow$4.0) & 48.6 ($\uparrow$18.8) & 52.4 ($\uparrow$6.8) & 57.6 ($\uparrow$11.3) & 62.1 ($\uparrow$5.8) \\
    \midrule
    \makecell[l]{Tarsier2-7B\\ \quad\textit{upgrading model}+\textit{data}} & 40.8 ($\uparrow$6.2) & 60.1 ($\uparrow$4.8) & 60.2 ($\uparrow$30.4) & 55.3 ($\uparrow$9.7) & 64.1 ($\uparrow$17.8) & 63.5 ($\uparrow$7.2)\\
    \bottomrule
    \end{tabular}
    }
    \caption{Results of the ablation study for pre-training. Tarsier1-7b-Qwen stands for the model where the base model is upgraded to Qwen2-VL, while the pre-training dataset remains the same as Tarsier1. Tarsier2 is trained from Qwen2-VL with an expanded pre-training dataset, growing from 13 million in Tarsier1 to 40 million samples.}
    \label{tab:pretrain_ab_vlm_and_data}
\end{table}

In this section, we evaluate the impact of several factors during pre-training, including the base model, pre-training data and training steps.
For the caption task, we report results after the SFT stage, which aligns the model's responses with the desired style. For other tasks, we report results after pre-training stage.

Compared to Tarsier1, two key improvements are made in the pre-training phase: upgrading the base model to Qwen2-VL and expanding the training dataset from 13 million to 40 million samples. Table~\ref{tab:pretrain_ab_vlm_and_data} illustrates the additive contributions for each improvement, showing that both enhancements consistently and significantly boost the model's performance in caption generation, video QA, and hallucination reduction. Specifically, these enhancements lead to accuracy improvements of 9.7\%, 17.8\%, and 7.2\% for short-video QA, long-video QA, and hallucination tests, respectively. For video description, the F1 score on the DREAM-1K dataset improves by 6.2\%.


\begin{figure}[t]
    \centering
    \includegraphics[width=0.7\linewidth]{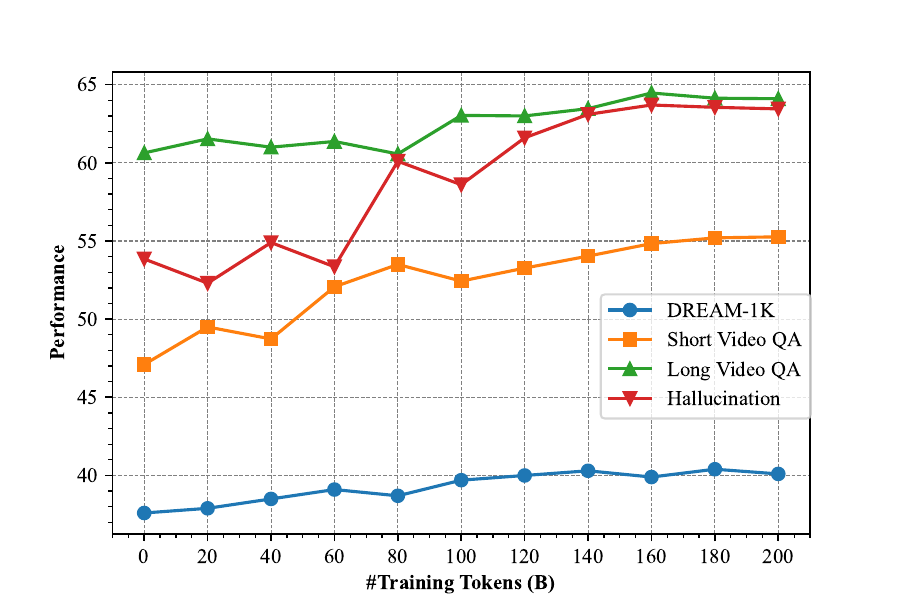}
    \caption{Model performance against training tokens. The results at the initial step reflect the performance of Qwen2-VL-7B.\protect\footnotemark}
    \label{fig:pretrain_ab_training_tokens}
\end{figure}

\footnotetext{For consistency across all checkpoints, we evaluate the Qwen2-VL-7B model using the same frame sampling strategy applied to other checkpoints. This may differ from the official sampling strategy in some benchmarks. For instance, the official setting of Video-MME uses 768 frames, while we sample 128 frames.}

To better understand the effect of the number of training tokens on pre-training performance, we plot the model’s performance as a function of token count during the pre-training stage, as shown in Figure~\ref{fig:pretrain_ab_training_tokens}. The results show that model performance improves with an increase in the number of training tokens, reaching convergence after 160 billion tokens. This suggests that a large volume of data is essential for optimal video understanding performance.


\subsubsection{SFT} 

\begin{table}[t]
    \centering
    \setlength{\tabcolsep}{2pt} 
    \resizebox{\textwidth}{!}{\begin{tabular}{l|ccc|cc|c}
    \toprule
    \multirow{2}{*}{\textbf{Model}} & \multicolumn{3}{c|}{\textbf{Caption}} & \multicolumn{2}{c|}{\textbf{Video QA}} & \multirow{2}{*}{\textbf{Hallucination}} \\
    & DREAM-1K & TempCompass-cg & Vinoground-Text & Short & Long & \\
    \midrule
    \modelname-7B-SFT & 40.8 & 60.1 & 60.2 & 56.2 & 63.2 & 71.9 \\
    \midrule
    \quad \textit{w/o SFT} & 35.2 ($\downarrow$5.6) & 50.5 ($\downarrow$9.6) & 57.2 ($\downarrow$3.0) & 55.3 ($\downarrow$0.9) & 64.1 ($\uparrow$0.9) & 63.5 ($\downarrow$8.4) \\
    \quad \textit{w/o grounding} & 37.4 ($\downarrow$3.4)  & 50.2 ($\downarrow$9.9)  & 60.6 ($\uparrow$0.4) & 55.9 ($\downarrow$0.3)  & 61.9 ($\downarrow$1.3) & 68.6 ($\downarrow$3.3)  \\
    \bottomrule
    \end{tabular}}
    \caption{Ablation study of temporal grounding dataset during the SFT phase. \modelname-7B-SFT refers to the model after the SFT phase. \textit{w/o SFT} refers to the model after pre-training; \textit{w/o grounding} refers to the model fine-tinued without grounding information.}
\label{tab:pretrain_ab_grounding}
\end{table}


The key factor in the SFT phase is fine-grained alignment. To investigate its impact, we conduct an ablation study, with the results presented in Table~\ref{tab:pretrain_ab_grounding}. When the video description data, which includes fine-grained temporal grounding information, is excluded (i.e., without grounding), model performance significantly deteriorates. Specifically, the F1 score on DREAM-1K decreases by 3.4\%, accuracy on TempCompass-cg drops by 9.9\%, accuracy on long-video QA falls by 1.3\%, and accuracy on the hallucination test declines by 3.3\%.

Furthermore, the SFT phase leads to substantial improvements, highlighting the importance of high-quality manually labeled data. It boosts the F1 score on DREAM-1K by 5.6\%, accuracy on TempCompass-cg by 9.6\%, accuracy on Vinoground-Text by 3.0\%, and accuracy on the hallucination test by 8.4\%, demonstrating the SFT phase's role in enhancing the model's fine-grained video understanding and mitigating hallucinations.

\subsubsection{DPO}

\begin{table}[t]
    \centering
    \scriptsize
    \setlength{\tabcolsep}{3pt} 
    \resizebox{\textwidth}{!}{\begin{tabular}{l|ccc|cc|c}
    \toprule
    \multirow{2}{*}{\textbf{Model}} & \multicolumn{3}{c|}{\textbf{Caption}} & \multicolumn{2}{c|}{\textbf{Video QA}} & \multirow{2}{*}{\textbf{Hallucination}} \\
    & DREAM-1K & TempCompass-cg & Vinoground-Text & Short & Long & \\
    \midrule
    Tarsier2-7B & 42.0 & 66.6 & 65.8 & 56.1 & 62.8 & 74.0  \\
    \midrule
    \quad \textit{w/o DPO} & 40.8 ($\downarrow$1.2) & 62.1 ($\downarrow$6.5) & 60.6 ($\downarrow$5.6) & 56.2 ($\uparrow$0.1) & 63.2 ($\uparrow$0.4) & 71.9 ($\downarrow$2.1) \\
    \quad \textit{w/o NS} & 41.5 ($\downarrow$0.5) & 61.1 ($\downarrow$5.5) & 59.8 ($\downarrow$6.0)& 56.1 ($\downarrow$0.0)  & 62.8 ($\downarrow$0.0) & 72.9 ($\downarrow$1.1) \\
    \quad \textit{w/o PF} & 40.5 ($\downarrow$1.5) & 65.1 ($\downarrow$1.5) & 67.6 ($\uparrow$1.8) & 56.0 ($\downarrow$0.1) & 62.3 ($\downarrow$0.5) & 74.2 ($\uparrow$0.2) \\
    \bottomrule
    \end{tabular}}
    \caption{Ablation study for DPO training phase, negative sampling (NS) and preference data filtering (PF) strategies.}
    \label{tab:dpo_ablation}
\end{table}

We conduct ablation experiments to evaluate the DPO phase, negative sampling (NS) and preference data filtering (PF) strategies. Specifically, we test the following settings: 1) {\bf \textit{w/o DPO}}: SFT model without DPO training. 2) {\bf \textit{w/o NS}}: Preference pairs generated by sampling the same video twice, without negative sampling. 3) {\bf \textit{w/o PF}}: Responses from negative sampling are treated as rejections, without utilizing AutoDQ Scorer to perform preference data filtering. For a fair comparison, the training data size and hyper-parameters for the latter two settings are kept consistent with the default setting, as detailed in Appendix~\ref{sec:dpo_ab_settings}.

As shown in Table~\ref{tab:dpo_ablation}, \modelname benefits a lot from the DPO training phase with significant improvement on caption tasks, especially TempCompass-cg (6.5\%) and Vinoground-Text (5.6\%). The hallucination capability also drops by 2.1\% without DPO, while the performance on video QA is not obviously affected. When further ablating dataset construction strategy of DPO, negative sampling plays an important role, without which the model results on most of the tasks are degraded to be almost the same as the SFT model (``\textit{w/o DPO}''), and the hallucination capability drops by 1.1\%. Additionally, preference data filtering with AutoDQ scorer has a significant impact on maintaining the quality of DPO datasets. As shown in Table~\ref{tab:dpo_ablation}, ``\textit{w/o PF}'' leads to degradation on more than a half of the tasks, and especially the DREAM-1K F1 score is even worse than the SFT model.

\subsection{Video Recaptioning using Tarsier2}
\begin{table}[t]
    \centering
    \scriptsize
    \setlength{\tabcolsep}{3pt} 
    \resizebox{\textwidth}{!}{\begin{tabular}{l|ccc|cc|c}
    \toprule
    \multirow{2}{*}{\textbf{Model}} & \multicolumn{3}{c|}{\textbf{Caption}} & \multicolumn{2}{c|}{\textbf{Video QA}} & \multirow{2}{*}{\textbf{Hallucination}} \\
    & DREAM-1K & TempCompass-cg & Vinoground-Text & Short & Long & \\
    \midrule
    Qwen2-VL-7B \cite{qwen2vl} & 31.2 & 54.2 & 40.0 & 49.4 & 60.3 & 51.9 \\
    \midrule
    \makecell[l]{+ \textit{Original FT}} & 35.2 ($\uparrow$4.0) & 49.9 ($\downarrow$4.3) & 39.0 ($\downarrow$1.0) & 46.9 ($\downarrow$2.5) & 55.4 ($\downarrow$4.9) & 43.0 ($\downarrow$8.9) \\
    \makecell[l]{+ \textit{Recaption FT}} & 39.5  ($\uparrow$8.3) & 67.7 ($\uparrow$13.5) & 55.0 ($\uparrow$15.0) & 52.5 ($\uparrow$3.1) & 56.8 ($\downarrow$3.5) & 68.5 ($\uparrow$16.6) \\
    \bottomrule
    \end{tabular}}
    \caption{The experimental results of recaptioning. ``\textit{Recaption FT}'' represents fine-tune the model on the Tarsier2-Recap-585K dataset. ``\textit{Original FT}'' represents fine-tune the model with the same videos as Tarsier2-Recap-585K but taking their original labels as target output.}
    \label{tab:recaption}
\end{table}
In this section, we utilize Tarsier2 as a captioner to generate detailed descriptions for a diverse set of 1M videos sourced from public datasets, resulting in the recaptioning dataset Tarsier2-Recap-585K\footnote{Tarsier2-Recap-585K is available on \href{https://huggingface.co/datasets/omni-research/Tarsier2-Recap-585K}{HuggingFace}.}. Details of the dataset composition are provided in Appendix \ref{sec:recaption_composition}.

We fine-tune Qwen2-VL-7B \cite{qwen2vl} on Tarsier2-Recap-585K and present the evaluation results in Table \ref{tab:recaption}. Fine-tuning on Tarsier2-Recap-585K significantly enhances the model's performance on detailed video description, achieving improvements in DREAM-1K (+8.3\%), TempCompass-cg (+13.4\%), and Vinoground-Text (+15.0\%). Moreover, it achieves an improvement of 16.6\% in hallucination test and an improvement of 3.1\% in short video-QA. 

In comparison, fine-tuning on the same 585K videos with original captions improves only the DREAM-1K F1 score (+4.0\%), while other metrics show significant declines. It indicates that the performance gains from Tarsier2-Recap-585K are primarily due to its high-quality and detailed captions rather than the additional training data volume.

Table \ref{tab:appendix-recap_detailed_results} in Appendix \ref{sec:ab_exp_detail} provides detailed benchmark results corresponding to Table \ref{tab:recaption}. These findings demonstrate that Tarsier2 can generate high-quality, detailed descriptions that offer fine-grained alignment information to help LVLMs to achieve significant improvements across various tasks.

\section{Conclusion}

In this paper, we introduce \modelname, a state-of-the-art large vision-language model that outperforms existing proprietary and open-source models in generating detailed and accurate video descriptions. Furthermore, \modelname sets new benchmarks across a wide range of video understanding tasks. Our ablation studies demonstrate that \modelname’s advancements are driven by scaling the volume and diversity of the training dataset, fine-grained temporal alignment, and DPO training.

Looking ahead, we outline several promising directions for future research. First, extending \modelname to handle longer video durations by developing more efficient model architectures and expanding the training dataset. Second, enhancing real-time video processing to improve the model’s ability to analyze and describe videos as they stream. Third, exploring richer interactions between video, audio, and text to create more comprehensive and context-aware video understanding systems.

\bibliographystyle{plain}
\bibliography{paper}

\begin{thebibliography}{100}

\bibitem{gpt4}
Josh Achiam, Steven Adler, Sandhini Agarwal, Lama Ahmad, Ilge Akkaya, Florencia~Leoni Aleman, Diogo Almeida, Janko Altenschmidt, Sam Altman, Shyamal Anadkat, et~al.
\newblock Gpt-4 technical report.
\newblock {\em arXiv preprint arXiv:2303.08774}, 2023.

\bibitem{agrawal2024pixtral}
Pravesh Agrawal, Szymon Antoniak, Emma~Bou Hanna, Baptiste Bout, Devendra Chaplot, Jessica Chudnovsky, Diogo Costa, Baudouin De~Monicault, Saurabh Garg, Theophile Gervet, et~al.
\newblock Pixtral 12b.
\newblock {\em arXiv preprint arXiv:2410.07073}, 2024.

\bibitem{an2025multi}
Xiang An, Kaicheng Yang, Xiangzi Dai, Ziyong Feng, and Jiankang Deng.
\newblock Multi-label cluster discrimination for visual representation learning.
\newblock In {\em European Conference on Computer Vision}, pages 428--444. Springer, 2025.

\bibitem{anne2017localizing}
Lisa Anne~Hendricks, Oliver Wang, Eli Shechtman, Josef Sivic, Trevor Darrell, and Bryan Russell.
\newblock Localizing moments in video with natural language.
\newblock In {\em Proceedings of the IEEE international conference on computer vision}, pages 5803--5812, 2017.

\bibitem{gpt4v}
Sally Applin, Gerardo Adesso, Rubaid Ashfaq, Max Bai, Matthew Brammer, Ethan Fecht, Andrew Goodman, Shelby Grossman, Matthew Groh, Hannah~Rose Kirk, et~al.
\newblock Gpt-4v (ision) system card.
\newblock 2023.

\bibitem{ataallah2024minigpt4}
Kirolos Ataallah, Xiaoqian Shen, Eslam Abdelrahman, Essam Sleiman, Deyao Zhu, Jian Ding, and Mohamed Elhoseiny.
\newblock Minigpt4-video: Advancing multimodal llms for video understanding with interleaved visual-textual tokens.
\newblock {\em arXiv preprint arXiv:2404.03413}, 2024.

\bibitem{awad2023trecvid}
George Awad, Keith Curtis, Asad Butt, Jonathan Fiscus, Afzal Godil, Yooyoung Lee, Andrew Delgado, Eliot Godard, Lukas Diduch, Jeffrey Liu, et~al.
\newblock An overview on the evaluated video retrieval tasks at trecvid 2022.
\newblock {\em arXiv preprint arXiv:2306.13118}, 2023.

\bibitem{bai2023qwen}
Jinze Bai, Shuai Bai, Shusheng Yang, Shijie Wang, Sinan Tan, Peng Wang, Junyang Lin, Chang Zhou, and Jingren Zhou.
\newblock Qwen-vl: A versatile vision-language model for understanding, localization, text reading, and beyond.
\newblock {\em arXiv preprint arXiv:2308.12966}, 1(2):3, 2023.

\bibitem{bai2024glance}
Ziyi Bai, Ruiping Wang, and Xilin Chen.
\newblock Glance and focus: Memory prompting for multi-event video question answering.
\newblock {\em Advances in Neural Information Processing Systems}, 36, 2024.

\bibitem{bain2021frozen}
Max Bain, Arsha Nagrani, G{\"u}l Varol, and Andrew Zisserman.
\newblock Frozen in time: A joint video and image encoder for end-to-end retrieval.
\newblock In {\em Proceedings of the IEEE/CVF International Conference on Computer Vision}, pages 1728--1738, 2021.

\bibitem{bhagavatula2019abductive}
Chandra Bhagavatula, Ronan~Le Bras, Chaitanya Malaviya, Keisuke Sakaguchi, Ari Holtzman, Hannah Rashkin, Doug Downey, Scott Wen-tau Yih, and Yejin Choi.
\newblock Abductive commonsense reasoning.
\newblock {\em arXiv preprint arXiv:1908.05739}, 2019.

\bibitem{cai2024temporalbench}
Mu~Cai, Reuben Tan, Jianrui Zhang, Bocheng Zou, Kai Zhang, Feng Yao, Fangrui Zhu, Jing Gu, Yiwu Zhong, Yuzhang Shang, et~al.
\newblock Temporalbench: Benchmarking fine-grained temporal understanding for multimodal video models.
\newblock {\em arXiv preprint arXiv:2410.10818}, 2024.

\bibitem{carreira2017quo}
Joao Carreira and Andrew Zisserman.
\newblock Quo vadis, action recognition? a new model and the kinetics dataset.
\newblock In {\em proceedings of the IEEE Conference on Computer Vision and Pattern Recognition}, pages 6299--6308, 2017.

\bibitem{chen-dolan-2011-collecting}
David Chen and William Dolan.
\newblock Collecting highly parallel data for paraphrase evaluation.
\newblock In Dekang Lin, Yuji Matsumoto, and Rada Mihalcea, editors, {\em Proceedings of the 49th Annual Meeting of the Association for Computational Linguistics: Human Language Technologies}, pages 190--200, Portland, Oregon, USA, June 2011. Association for Computational Linguistics.

\bibitem{chen2023sharegpt4v}
Lin Chen, Jisong Li, Xiaoyi Dong, Pan Zhang, Conghui He, Jiaqi Wang, Feng Zhao, and Dahua Lin.
\newblock Sharegpt4v: Improving large multi-modal models with better captions.
\newblock {\em arXiv preprint arXiv:2311.12793}, 2023.

\bibitem{chen2024sharegpt4video}
Lin Chen, Xilin Wei, Jinsong Li, Xiaoyi Dong, Pan Zhang, Yuhang Zang, Zehui Chen, Haodong Duan, Bin Lin, Zhenyu Tang, Li~Yuan, Yu~Qiao, Dahua Lin, Feng Zhao, and Jiaqi Wang.
\newblock Sharegpt4video: Improving video understanding and generation with better captions, 2024.

\bibitem{chen2023vast}
Sihan Chen, Handong Li, Qunbo Wang, Zijia Zhao, Mingzhen Sun, Xinxin Zhu, and Jing Liu.
\newblock Vast: A vision-audio-subtitle-text omni-modality foundation model and dataset.
\newblock {\em Advances in Neural Information Processing Systems}, 36:72842--72866, 2023.

\bibitem{chen2024panda}
Tsai-Shien Chen, Aliaksandr Siarohin, Willi Menapace, Ekaterina Deyneka, Hsiang-wei Chao, Byung~Eun Jeon, Yuwei Fang, Hsin-Ying Lee, Jian Ren, Ming-Hsuan Yang, et~al.
\newblock Panda-70m: Captioning 70m videos with multiple cross-modality teachers.
\newblock In {\em Proceedings of the IEEE/CVF Conference on Computer Vision and Pattern Recognition}, pages 13320--13331, 2024.

\bibitem{chen2023autoeval}
Xiuyuan Chen, Yuan Lin, Yuchen Zhang, and Weiran Huang.
\newblock Autoeval-video: An automatic benchmark for assessing large vision language models in open-ended video question answering.
\newblock {\em arXiv preprint arXiv:2311.14906}, 2023.

\bibitem{chen2024expanding}
Zhe Chen, Weiyun Wang, Yue Cao, Yangzhou Liu, Zhangwei Gao, Erfei Cui, Jinguo Zhu, Shenglong Ye, Hao Tian, Zhaoyang Liu, et~al.
\newblock Expanding performance boundaries of open-source multimodal models with model, data, and test-time scaling.
\newblock {\em arXiv preprint arXiv:2412.05271}, 2024.

\bibitem{chen2024far}
Zhe Chen, Weiyun Wang, Hao Tian, Shenglong Ye, Zhangwei Gao, Erfei Cui, Wenwen Tong, Kongzhi Hu, Jiapeng Luo, Zheng Ma, et~al.
\newblock How far are we to gpt-4v? closing the gap to commercial multimodal models with open-source suites.
\newblock {\em arXiv preprint arXiv:2404.16821}, 2024.

\bibitem{chen2024internvl}
Zhe Chen, Jiannan Wu, Wenhai Wang, Weijie Su, Guo Chen, Sen Xing, Muyan Zhong, Qinglong Zhang, Xizhou Zhu, Lewei Lu, et~al.
\newblock Internvl: Scaling up vision foundation models and aligning for generic visual-linguistic tasks.
\newblock In {\em Proceedings of the IEEE/CVF Conference on Computer Vision and Pattern Recognition}, pages 24185--24198, 2024.

\bibitem{cheng2024videollama2}
Zesen Cheng, Sicong Leng, Hang Zhang, Yifei Xin, Xin Li, Guanzheng Chen, Yongxin Zhu, Wenqi Zhang, Ziyang Luo, Deli Zhao, and Lidong Bing.
\newblock Videollama 2: Advancing spatial-temporal modeling and audio understanding in video-llms, 2024.

\bibitem{vicuna2023}
Wei-Lin Chiang, Zhuohan Li, Zi~Lin, Ying Sheng, Zhanghao Wu, Hao Zhang, Lianmin Zheng, Siyuan Zhuang, Yonghao Zhuang, Joseph~E. Gonzalez, Ion Stoica, and Eric~P. Xing.
\newblock Vicuna: An open-source chatbot impressing gpt-4 with 90\%* chatgpt quality, March 2023.

\bibitem{cores2024tvbench}
Daniel Cores, Michael Dorkenwald, Manuel Mucientes, Cees~GM Snoek, and Yuki~M Asano.
\newblock Tvbench: Redesigning video-language evaluation.
\newblock {\em arXiv preprint arXiv:2410.07752}, 2024.

\bibitem{sharegpt4o}
Erfei Cui, Yinan He, Zheng Ma, Zhe Chen, Hao Tian, Weiyun Wang, Kunchang Li, Yi~Wang, Wenhai Wang, Xizhou Zhu, Lewei Lu, Tong Lu, Yali Wang, Limin Wang, Yu~Qiao, and Jifeng Dai.
\newblock Sharegpt-4o: Comprehensive multimodal annotations with gpt-4o.
\newblock \url{https://sharegpt4o.github.io/}, 2024.

\bibitem{dubey2024llama}
Abhimanyu Dubey, Abhinav Jauhri, Abhinav Pandey, Abhishek Kadian, Ahmad Al-Dahle, Aiesha Letman, Akhil Mathur, Alan Schelten, Amy Yang, Angela Fan, et~al.
\newblock The llama 3 herd of models.
\newblock {\em arXiv preprint arXiv:2407.21783}, 2024.

\bibitem{epstein2020oops}
Dave Epstein, Boyuan Chen, and Carl Vondrick.
\newblock Oops! predicting unintentional action in video.
\newblock In {\em Proceedings of the IEEE/CVF conference on computer vision and pattern recognition}, pages 919--929, 2020.

\bibitem{fan2019egovqa}
Chenyou Fan.
\newblock Egovqa-an egocentric video question answering benchmark dataset.
\newblock In {\em Proceedings of the IEEE/CVF International Conference on Computer Vision Workshops}, pages 0--0, 2019.

\bibitem{fang2024mmbench}
Xinyu Fang, Kangrui Mao, Haodong Duan, Xiangyu Zhao, Yining Li, Dahua Lin, and Kai Chen.
\newblock Mmbench-video: A long-form multi-shot benchmark for holistic video understanding.
\newblock {\em arXiv preprint arXiv:2406.14515}, 2024.

\bibitem{fu2024video}
Chaoyou Fu, Yuhan Dai, Yongdong Luo, Lei Li, Shuhuai Ren, Renrui Zhang, Zihan Wang, Chenyu Zhou, Yunhang Shen, Mengdan Zhang, et~al.
\newblock Video-mme: The first-ever comprehensive evaluation benchmark of multi-modal llms in video analysis.
\newblock {\em arXiv preprint arXiv:2405.21075}, 2024.

\bibitem{gao2017tall}
Jiyang Gao, Chen Sun, Zhenheng Yang, and Ram Nevatia.
\newblock Tall: Temporal activity localization via language query.
\newblock In {\em Proceedings of the IEEE international conference on computer vision}, pages 5267--5275, 2017.

\bibitem{gao2023llama}
Peng Gao, Jiaming Han, Renrui Zhang, Ziyi Lin, Shijie Geng, Aojun Zhou, Wei Zhang, Pan Lu, Conghui He, Xiangyu Yue, et~al.
\newblock Llama-adapter v2: Parameter-efficient visual instruction model.
\newblock {\em arXiv preprint arXiv:2304.15010}, 2023.

\bibitem{goyal2017something}
Raghav Goyal, Samira Ebrahimi~Kahou, Vincent Michalski, Joanna Materzynska, Susanne Westphal, Heuna Kim, Valentin Haenel, Ingo Fruend, Peter Yianilos, Moritz Mueller-Freitag, et~al.
\newblock The" something something" video database for learning and evaluating visual common sense.
\newblock In {\em Proceedings of the IEEE international conference on computer vision}, pages 5842--5850, 2017.

\bibitem{grauman2022ego4d}
Kristen Grauman, Andrew Westbury, Eugene Byrne, Zachary Chavis, Antonino Furnari, Rohit Girdhar, Jackson Hamburger, Hao Jiang, Miao Liu, Xingyu Liu, et~al.
\newblock Ego4d: Around the world in 3,000 hours of egocentric video.
\newblock In {\em Proceedings of the IEEE/CVF Conference on Computer Vision and Pattern Recognition}, pages 18995--19012, 2022.

\bibitem{gu2018ava}
Chunhui Gu, Chen Sun, David~A Ross, Carl Vondrick, Caroline Pantofaru, Yeqing Li, Sudheendra Vijayanarasimhan, George Toderici, Susanna Ricco, Rahul Sukthankar, et~al.
\newblock Ava: A video dataset of spatio-temporally localized atomic visual actions.
\newblock In {\em Proceedings of the IEEE conference on computer vision and pattern recognition}, pages 6047--6056, 2018.

\bibitem{guo2024vtg}
Yongxin Guo, Jingyu Liu, Mingda Li, Xiaoying Tang, Xi~Chen, and Bo~Zhao.
\newblock Vtg-llm: Integrating timestamp knowledge into video llms for enhanced video temporal grounding.
\newblock {\em arXiv preprint arXiv:2405.13382}, 2024.

\bibitem{he2024storyteller}
Yichen He, Yuan Lin, Jianchao Wu, Hanchong Zhang, Yuchen Zhang, and Ruicheng Le.
\newblock Storyteller: Improving long video description through global audio-visual character identification.
\newblock {\em arXiv preprint arXiv:2411.07076}, 2024.

\bibitem{huang2024lita}
De-An Huang, Shijia Liao, Subhashree Radhakrishnan, Hongxu Yin, Pavlo Molchanov, Zhiding Yu, and Jan Kautz.
\newblock Lita: Language instructed temporal-localization assistant.
\newblock In {\em ECCV}, 2024.

\bibitem{huang2016visual}
Ting-Hao Huang, Francis Ferraro, Nasrin Mostafazadeh, Ishan Misra, Aishwarya Agrawal, Jacob Devlin, Ross Girshick, Xiaodong He, Pushmeet Kohli, Dhruv Batra, et~al.
\newblock Visual storytelling.
\newblock In {\em Proceedings of the 2016 conference of the North American chapter of the association for computational linguistics: Human language technologies}, pages 1233--1239, 2016.

\bibitem{gpt4o}
Aaron Hurst, Adam Lerer, Adam~P Goucher, Adam Perelman, Aditya Ramesh, Aidan Clark, AJ~Ostrow, Akila Welihinda, Alan Hayes, Alec Radford, et~al.
\newblock Gpt-4o system card.
\newblock {\em arXiv preprint arXiv:2410.21276}, 2024.

\bibitem{idrees2017thumos}
Haroon Idrees, Amir~R Zamir, Yu-Gang Jiang, Alex Gorban, Ivan Laptev, Rahul Sukthankar, and Mubarak Shah.
\newblock The thumos challenge on action recognition for videos “in the wild”.
\newblock {\em Computer Vision and Image Understanding}, 155:1--23, 2017.

\bibitem{jang2017tgif}
Yunseok Jang, Yale Song, Youngjae Yu, Youngjin Kim, and Gunhee Kim.
\newblock Tgif-qa: Toward spatio-temporal reasoning in visual question answering.
\newblock In {\em Proceedings of the IEEE conference on computer vision and pattern recognition}, pages 2758--2766, 2017.

\bibitem{jia2022egotaskqa}
Baoxiong Jia, Ting Lei, Song-Chun Zhu, and Siyuan Huang.
\newblock Egotaskqa: Understanding human tasks in egocentric videos.
\newblock In {\em The 36th Conference on Neural Information Processing Systems (NeurIPS 2022) Track on Datasets and Benchmarks}, 2022.

\bibitem{kenton2019bert}
Jacob Devlin Ming-Wei~Chang Kenton and Lee~Kristina Toutanova.
\newblock Bert: Pre-training of deep bidirectional transformers for language understanding.
\newblock In {\em Proceedings of naacL-HLT}, volume~1, page~2. Minneapolis, Minnesota, 2019.

\bibitem{kim2022donut}
Geewook Kim, Teakgyu Hong, Moonbin Yim, JeongYeon Nam, Jinyoung Park, Jinyeong Yim, Wonseok Hwang, Sangdoo Yun, Dongyoon Han, and Seunghyun Park.
\newblock Ocr-free document understanding transformer.
\newblock In {\em European Conference on Computer Vision (ECCV)}, 2022.

\bibitem{krishna2017dense}
Ranjay Krishna, Kenji Hata, Frederic Ren, Li~Fei-Fei, and Juan~Carlos Niebles.
\newblock Dense-captioning events in videos.
\newblock In {\em International Conference on Computer Vision (ICCV)}, 2017.

\bibitem{krishna2017visual}
Ranjay Krishna, Yuke Zhu, Oliver Groth, Justin Johnson, Kenji Hata, Joshua Kravitz, Stephanie Chen, Yannis Kalantidis, Li-Jia Li, David~A Shamma, et~al.
\newblock Visual genome: Connecting language and vision using crowdsourced dense image annotations.
\newblock {\em International journal of computer vision}, 123:32--73, 2017.

\bibitem{kuehne2011hmdb}
Hildegard Kuehne, Hueihan Jhuang, Est{\'\i}baliz Garrote, Tomaso Poggio, and Thomas Serre.
\newblock Hmdb: a large video database for human motion recognition.
\newblock In {\em 2011 International conference on computer vision}, pages 2556--2563. IEEE, 2011.

\bibitem{le2020hierarchical}
Thao~Minh Le, Vuong Le, Svetha Venkatesh, and Truyen Tran.
\newblock Hierarchical conditional relation networks for video question answering.
\newblock In {\em Proceedings of the IEEE/CVF conference on computer vision and pattern recognition}, pages 9972--9981, 2020.

\bibitem{lei2021detecting}
Jie Lei, Tamara~L Berg, and Mohit Bansal.
\newblock Detecting moments and highlights in videos via natural language queries.
\newblock {\em Advances in Neural Information Processing Systems}, 34:11846--11858, 2021.

\bibitem{li2024llava}
Bo~Li, Hao Zhang, Kaichen Zhang, Dong Guo, Yuanhan Zhang, Renrui Zhang, Feng Li, Ziwei Liu, and Chunyuan Li.
\newblock Llava-next: What else influences visual instruction tuning beyond data, 2024.

\bibitem{li2024llavanext}
Bo~Li, Yuanhan Zhang, Dong Guo, Renrui Zhang, Feng Li, Hao Zhang, Kaichen Zhang, Peiyuan Zhang, Yanwei Li, Ziwei Liu, et~al.
\newblock Llava-onevision: Easy visual task transfer.
\newblock {\em arXiv preprint arXiv:2408.03326}, 2024.

\bibitem{li2024aria}
Dongxu Li, Yudong Liu, Haoning Wu, Yue Wang, Zhiqi Shen, Bowen Qu, Xinyao Niu, Guoyin Wang, Bei Chen, and Junnan Li.
\newblock Aria: An open multimodal native mixture-of-experts model.
\newblock {\em arXiv preprint arXiv:2410.05993}, 2024.

\bibitem{li2023blip}
Junnan Li, Dongxu Li, Silvio Savarese, and Steven Hoi.
\newblock Blip-2: Bootstrapping language-image pre-training with frozen image encoders and large language models.
\newblock In {\em International conference on machine learning}, pages 19730--19742. PMLR, 2023.

\bibitem{li2023videochat}
KunChang Li, Yinan He, Yi~Wang, Yizhuo Li, Wenhai Wang, Ping Luo, Yali Wang, Limin Wang, and Yu~Qiao.
\newblock Videochat: Chat-centric video understanding.
\newblock {\em arXiv preprint arXiv:2305.06355}, 2023.

\bibitem{li2024mvbench}
Kunchang Li, Yali Wang, Yinan He, Yizhuo Li, Yi~Wang, Yi~Liu, Zun Wang, Jilan Xu, Guo Chen, Ping Luo, et~al.
\newblock Mvbench: A comprehensive multi-modal video understanding benchmark.
\newblock In {\em Proceedings of the IEEE/CVF Conference on Computer Vision and Pattern Recognition}, pages 22195--22206, 2024.

\bibitem{li2025llama}
Yanwei Li, Chengyao Wang, and Jiaya Jia.
\newblock Llama-vid: An image is worth 2 tokens in large language models.
\newblock In {\em European Conference on Computer Vision}, pages 323--340. Springer, 2025.

\bibitem{li2016tgif}
Yuncheng Li, Yale Song, Liangliang Cao, Joel Tetreault, Larry Goldberg, Alejandro Jaimes, and Jiebo Luo.
\newblock Tgif: A new dataset and benchmark on animated gif description.
\newblock In {\em Proceedings of the IEEE Conference on Computer Vision and Pattern Recognition}, pages 4641--4650, 2016.

\bibitem{lian2023openorca}
W~Lian, B~Goodson, E~Pentland, et~al.
\newblock Openorca: An open dataset of gpt augmented flan reasoning traces, 2023.

\bibitem{lin2023video}
Bin Lin, Bin Zhu, Yang Ye, Munan Ning, Peng Jin, and Li~Yuan.
\newblock Video-llava: Learning united visual representation by alignment before projection.
\newblock {\em arXiv preprint arXiv:2311.10122}, 2023.

\bibitem{lin2024vila}
Ji~Lin, Hongxu Yin, Wei Ping, Pavlo Molchanov, Mohammad Shoeybi, and Song Han.
\newblock Vila: On pre-training for visual language models.
\newblock In {\em Proceedings of the IEEE/CVF Conference on Computer Vision and Pattern Recognition}, pages 26689--26699, 2024.

\bibitem{lin2014microsoft}
Tsung-Yi Lin, Michael Maire, Serge Belongie, James Hays, Pietro Perona, Deva Ramanan, Piotr Doll{\'a}r, and C~Lawrence Zitnick.
\newblock Microsoft coco: Common objects in context.
\newblock In {\em Computer Vision--ECCV 2014: 13th European Conference, Zurich, Switzerland, September 6-12, 2014, Proceedings, Part V 13}, pages 740--755. Springer, 2014.

\bibitem{liu2023improved}
Haotian Liu, Chunyuan Li, Yuheng Li, and Yong~Jae Lee.
\newblock Improved baselines with visual instruction tuning.
\newblock {\em arXiv preprint arXiv:2310.03744}, 2023.

\bibitem{liu2024kangaroo}
Jiajun Liu, Yibing Wang, Hanghang Ma, Xiaoping Wu, Xiaoqi Ma, Xiaoming Wei, Jianbin Jiao, Enhua Wu, and Jie Hu.
\newblock Kangaroo: A powerful video-language model supporting long-context video input.
\newblock {\em arXiv preprint arXiv:2408.15542}, 2024.

\bibitem{liu2024robomamba}
Jiaming Liu, Mengzhen Liu, Zhenyu Wang, Lily Lee, Kaichen Zhou, Pengju An, Senqiao Yang, Renrui Zhang, Yandong Guo, and Shanghang Zhang.
\newblock Robomamba: Multimodal state space model for efficient robot reasoning and manipulation.
\newblock {\em arXiv preprint arXiv:2406.04339}, 2024.

\bibitem{liu2024etbench}
Ye~Liu, Zongyang Ma, Zhongang Qi, Yang Wu, Chang~Wen Chen, and Ying Shan.
\newblock E.t. bench: Towards open-ended event-level video-language understanding.
\newblock In {\em Neural Information Processing Systems (NeurIPS)}, 2024.

\bibitem{liu2022fineaction}
Yi~Liu, Limin Wang, Yali Wang, Xiao Ma, and Yu~Qiao.
\newblock Fineaction: A fine-grained video dataset for temporal action localization.
\newblock {\em IEEE transactions on image processing}, 31:6937--6950, 2022.

\bibitem{liu2024tempcompass}
Yuanxin Liu, Shicheng Li, Yi~Liu, Yuxiang Wang, Shuhuai Ren, Lei Li, Sishuo Chen, Xu~Sun, and Lu~Hou.
\newblock Tempcompass: Do video llms really understand videos?
\newblock {\em arXiv preprint arXiv:2403.00476}, 2024.

\bibitem{liu2024nvila}
Zhijian Liu, Ligeng Zhu, Baifeng Shi, Zhuoyang Zhang, Yuming Lou, Shang Yang, Haocheng Xi, Shiyi Cao, Yuxian Gu, Dacheng Li, Xiuyu Li, Yunhao Fang, Yukang Chen, Cheng-Yu Hsieh, De-An Huang, An-Chieh Cheng, Vishwesh Nath, Jinyi Hu, Sifei Liu, Ranjay Krishna, Daguang Xu, Xiaolong Wang, Pavlo Molchanov, Jan Kautz, Hongxu Yin, Song Han, and Yao Lu.
\newblock Nvila: Efficient frontier visual language models, 2024.

\bibitem{liu2024mmdu}
Ziyu Liu, Tao Chu, Yuhang Zang, Xilin Wei, Xiaoyi Dong, Pan Zhang, Zijian Liang, Yuanjun Xiong, Yu~Qiao, Dahua Lin, et~al.
\newblock Mmdu: A multi-turn multi-image dialog understanding benchmark and instruction-tuning dataset for lvlms.
\newblock {\em arXiv preprint arXiv:2406.11833}, 2024.

\bibitem{liu2024oryx}
Zuyan Liu, Yuhao Dong, Ziwei Liu, Winston Hu, Jiwen Lu, and Yongming Rao.
\newblock Oryx mllm: On-demand spatial-temporal understanding at arbitrary resolution.
\newblock {\em arXiv preprint arXiv:2409.12961}, 2024.

\bibitem{long2022towards}
Shangbang Long, Siyang Qin, Dmitry Panteleev, Alessandro Bissacco, Yasuhisa Fujii, and Michalis Raptis.
\newblock Towards end-to-end unified scene text detection and layout analysis.
\newblock In {\em Proceedings of the IEEE/CVF Conference on Computer Vision and Pattern Recognition}, 2022.

\bibitem{lucas2005icdar}
Simon~M Lucas, Alex Panaretos, Luis Sosa, Anthony Tang, Shirley Wong, Robert Young, Kazuki Ashida, Hiroki Nagai, Masayuki Okamoto, Hiroaki Yamamoto, et~al.
\newblock Icdar 2003 robust reading competitions: entries, results, and future directions.
\newblock {\em International Journal of Document Analysis and Recognition (IJDAR)}, 7:105--122, 2005.

\bibitem{luo2023valley}
Ruipu Luo, Ziwang Zhao, Min Yang, Junwei Dong, Minghui Qiu, Pengcheng Lu, Tao Wang, and Zhongyu Wei.
\newblock Valley: Video assistant with large language model enhanced ability.
\newblock {\em arXiv preprint arXiv:2306.07207}, 2023.

\bibitem{maaz2023video}
Muhammad Maaz, Hanoona Rasheed, Salman Khan, and Fahad~Shahbaz Khan.
\newblock Video-chatgpt: Towards detailed video understanding via large vision and language models.
\newblock {\em arXiv preprint arXiv:2306.05424}, 2023.

\bibitem{majumdar2024openeqa}
Arjun Majumdar, Anurag Ajay, Xiaohan Zhang, Pranav Putta, Sriram Yenamandra, Mikael Henaff, Sneha Silwal, Paul Mcvay, Oleksandr Maksymets, Sergio Arnaud, et~al.
\newblock Openeqa: Embodied question answering in the era of foundation models.
\newblock In {\em Proceedings of the IEEE/CVF Conference on Computer Vision and Pattern Recognition}, pages 16488--16498, 2024.

\bibitem{mangalam2023egoschema}
Karttikeya Mangalam, Raiymbek Akshulakov, and Jitendra Malik.
\newblock Egoschema: A diagnostic benchmark for very long-form video language understanding.
\newblock {\em Advances in Neural Information Processing Systems}, 36:46212--46244, 2023.

\bibitem{materzynska2019jester}
Joanna Materzynska, Guillaume Berger, Ingo Bax, and Roland Memisevic.
\newblock The jester dataset: A large-scale video dataset of human gestures.
\newblock In {\em Proceedings of the IEEE/CVF international conference on computer vision workshops}, pages 0--0, 2019.

\bibitem{miech2020rareact}
Antoine Miech, Jean-Baptiste Alayrac, Ivan Laptev, Josef Sivic, and Andrew Zisserman.
\newblock Rareact: A video dataset of unusual interactions.
\newblock {\em arXiv preprint arXiv:2008.01018}, 2020.

\bibitem{miech2019howto100m}
Antoine Miech, Dimitri Zhukov, Jean-Baptiste Alayrac, Makarand Tapaswi, Ivan Laptev, and Josef Sivic.
\newblock Howto100m: Learning a text-video embedding by watching hundred million narrated video clips.
\newblock In {\em Proceedings of the IEEE/CVF international conference on computer vision}, pages 2630--2640, 2019.

\bibitem{monfort2021spoken}
Mathew Monfort, SouYoung Jin, Alexander Liu, David Harwath, Rogerio Feris, James Glass, and Aude Oliva.
\newblock Spoken moments: Learning joint audio-visual representations from video descriptions.
\newblock In {\em Proceedings of the IEEE/CVF Conference on Computer Vision and Pattern Recognition}, pages 14871--14881, 2021.

\bibitem{monfort2021multi}
Mathew Monfort, Bowen Pan, Kandan Ramakrishnan, Alex Andonian, Barry~A McNamara, Alex Lascelles, Quanfu Fan, Dan Gutfreund, Rog{\'e}rio~Schmidt Feris, and Aude Oliva.
\newblock Multi-moments in time: Learning and interpreting models for multi-action video understanding.
\newblock {\em IEEE Transactions on Pattern Analysis and Machine Intelligence}, 44(12):9434--9445, 2021.

\bibitem{ordonez2011im2text}
Vicente Ordonez, Girish Kulkarni, and Tamara Berg.
\newblock Im2text: Describing images using 1 million captioned photographs.
\newblock {\em Advances in neural information processing systems}, 24, 2011.

\bibitem{park2019cord}
Seunghyun Park, Seung Shin, Bado Lee, Junyeop Lee, Jaeheung Surh, Minjoon Seo, and Hwalsuk Lee.
\newblock Cord: a consolidated receipt dataset for post-ocr parsing.
\newblock In {\em Workshop on Document Intelligence at NeurIPS 2019}, 2019.

\bibitem{patraucean2024perception}
Viorica Patraucean, Lucas Smaira, Ankush Gupta, Adria Recasens, Larisa Markeeva, Dylan Banarse, Skanda Koppula, Mateusz Malinowski, Yi~Yang, Carl Doersch, et~al.
\newblock Perception test: A diagnostic benchmark for multimodal video models.
\newblock {\em Advances in Neural Information Processing Systems}, 36, 2024.

\bibitem{plummer2015flickr30k}
Bryan~A Plummer, Liwei Wang, Chris~M Cervantes, Juan~C Caicedo, Julia Hockenmaier, and Svetlana Lazebnik.
\newblock Flickr30k entities: Collecting region-to-phrase correspondences for richer image-to-sentence models.
\newblock In {\em Proceedings of the IEEE international conference on computer vision}, pages 2641--2649, 2015.

\bibitem{pramanick2023egovlpv2}
Shraman Pramanick, Yale Song, Sayan Nag, Kevin~Qinghong Lin, Hardik Shah, Mike~Zheng Shou, Rama Chellappa, and Pengchuan Zhang.
\newblock Egovlpv2: Egocentric video-language pre-training with fusion in the backbone.
\newblock In {\em Proceedings of the IEEE/CVF International Conference on Computer Vision}, pages 5285--5297, 2023.

\bibitem{rafailov2024direct}
Rafael Rafailov, Archit Sharma, Eric Mitchell, Christopher~D Manning, Stefano Ermon, and Chelsea Finn.
\newblock Direct preference optimization: Your language model is secretly a reward model.
\newblock {\em Advances in Neural Information Processing Systems}, 36, 2024.

\bibitem{regneri2013grounding}
Michaela Regneri, Marcus Rohrbach, Dominikus Wetzel, Stefan Thater, Bernt Schiele, and Manfred Pinkal.
\newblock Grounding action descriptions in videos.
\newblock {\em Transactions of the Association for Computational Linguistics}, 1:25--36, 2013.

\bibitem{Ren2023TimeChat}
Shuhuai Ren, Linli Yao, Shicheng Li, Xu~Sun, and Lu~Hou.
\newblock Timechat: A time-sensitive multimodal large language model for long video understanding.
\newblock {\em ArXiv}, abs/2312.02051, 2023.

\bibitem{rohrbach2017movie}
Anna Rohrbach, Atousa Torabi, Marcus Rohrbach, Niket Tandon, Christopher Pal, Hugo Larochelle, Aaron Courville, and Bernt Schiele.
\newblock Movie description.
\newblock {\em International Journal of Computer Vision}, 123:94--120, 2017.

\bibitem{robovqa2023arxiv}
Pierre Sermanet, Tianli Ding, Jeffrey Zhao, Fei Xia, Debidatta Dwibedi, Keerthana Gopalakrishnan, Christine Chan, Gabriel Dulac-Arnold, Sharath Maddineni, Nikhil~J Joshi, Pete Florence, Wei Han, Robert Baruch, Yao Lu, Suvir Mirchandani, Peng Xu, Pannag Sanketi, Karol Hausman, Izhak Shafran, Brian Ichter, and Yuan Cao.
\newblock Robovqa: Multimodal long-horizon reasoning for robotics.
\newblock In {\em arXiv preprint arXiv:2311.00899}, 2023.

\bibitem{shangguan2024tomato}
Ziyao Shangguan, Chuhan Li, Yuxuan Ding, Yanan Zheng, Yilun Zhao, Tesca Fitzgerald, and Arman Cohan.
\newblock Tomato: Assessing visual temporal reasoning capabilities in multimodal foundation models.
\newblock {\em arXiv preprint arXiv:2410.23266}, 2024.

\bibitem{shi2017icdar2017}
Baoguang Shi, Cong Yao, Minghui Liao, Mingkun Yang, Pei Xu, Linyan Cui, Serge Belongie, Shijian Lu, and Xiang Bai.
\newblock Icdar2017 competition on reading chinese text in the wild (rctw-17).
\newblock In {\em 2017 14th iapr international conference on document analysis and recognition (ICDAR)}, volume~1, pages 1429--1434. IEEE, 2017.

\bibitem{sigurdsson2018charades}
Gunnar~A Sigurdsson, Abhinav Gupta, Cordelia Schmid, Ali Farhadi, and Karteek Alahari.
\newblock Charades-ego: A large-scale dataset of paired third and first person videos.
\newblock {\em arXiv preprint arXiv:1804.09626}, 2018.

\bibitem{sigurdsson2016hollywood}
Gunnar~A Sigurdsson, G{\"u}l Varol, Xiaolong Wang, Ali Farhadi, Ivan Laptev, and Abhinav Gupta.
\newblock Hollywood in homes: Crowdsourcing data collection for activity understanding.
\newblock In {\em Computer Vision--ECCV 2016: 14th European Conference, Amsterdam, The Netherlands, October 11--14, 2016, Proceedings, Part I 14}, pages 510--526. Springer, 2016.

\bibitem{sun2019icdar}
Yipeng Sun, Zihan Ni, Chee-Kheng Chng, Yuliang Liu, Canjie Luo, Chun~Chet Ng, Junyu Han, Errui Ding, Jingtuo Liu, Dimosthenis Karatzas, et~al.
\newblock Icdar 2019 competition on large-scale street view text with partial labeling-rrc-lsvt.
\newblock In {\em 2019 International Conference on Document Analysis and Recognition (ICDAR)}, pages 1557--1562. IEEE, 2019.

\bibitem{tang2024enhancing}
Changli Tang, Yixuan Li, Yudong Yang, Jimin Zhuang, Guangzhi Sun, Wei Li, Zujun Ma, and Chao Zhang.
\newblock Enhancing multimodal llm for detailed and accurate video captioning using multi-round preference optimization, 2024.

\bibitem{tang2024enhancingmultimodalllmdetailed}
Changli Tang, Yixuan Li, Yudong Yang, Jimin Zhuang, Guangzhi Sun, Wei Li, Zujun Ma, and Chao Zhang.
\newblock Enhancing multimodal llm for detailed and accurate video captioning using multi-round preference optimization, 2024.

\bibitem{tang2019coin}
Yansong Tang, Dajun Ding, Yongming Rao, Yu~Zheng, Danyang Zhang, Lili Zhao, Jiwen Lu, and Jie Zhou.
\newblock Coin: A large-scale dataset for comprehensive instructional video analysis.
\newblock In {\em Proceedings of the IEEE/CVF Conference on Computer Vision and Pattern Recognition}, pages 1207--1216, 2019.

\bibitem{geminiteam2024gemini15unlockingmultimodal}
Gemini Team, Petko Georgiev, Ving~Ian Lei, Ryan Burnell, Libin Bai, Anmol Gulati, Garrett Tanzer, Damien Vincent, Zhufeng Pan, Shibo Wang, Soroosh Mariooryad, Yifan Ding, et~al.
\newblock Gemini 1.5: Unlocking multimodal understanding across millions of tokens of context, 2024.

\bibitem{veit2016coco}
Andreas Veit, Tomas Matera, Lukas Neumann, Jiri Matas, and Serge Belongie.
\newblock Coco-text: Dataset and benchmark for text detection and recognition in natural images.
\newblock {\em arXiv preprint arXiv:1601.07140}, 2016.

\bibitem{wang2024elysium}
Han Wang, Yanjie Wang, Yongjie Ye, Yuxiang Nie, and Can Huang.
\newblock Elysium: Exploring object-level perception in videos via mllm.
\newblock {\em arXiv preprint arXiv:2403.16558}, 2024.

\bibitem{wang2024tarsierrecipestrainingevaluating}
Jiawei Wang, Liping Yuan, Yuchen Zhang, and Haomiao Sun.
\newblock Tarsier: Recipes for training and evaluating large video description models, 2024.

\bibitem{qwen2vl}
Peng Wang, Shuai Bai, Sinan Tan, Shijie Wang, Zhihao Fan, Jinze Bai, Keqin Chen, Xuejing Liu, Jialin Wang, Wenbin Ge, Yang Fan, Kai Dang, Mengfei Du, Xuancheng Ren, Rui Men, Dayiheng Liu, Chang Zhou, Jingren Zhou, and Junyang Lin.
\newblock Qwen2-vl: Enhancing vision-language model's perception of the world at any resolution.
\newblock {\em arXiv preprint arXiv:2409.12191}, 2024.

\bibitem{wang2019vatex}
Xin Wang, Jiawei Wu, Junkun Chen, Lei Li, Yuan-Fang Wang, and William~Yang Wang.
\newblock Vatex: A large-scale, high-quality multilingual dataset for video-and-language research.
\newblock In {\em Proceedings of the IEEE/CVF international conference on computer vision}, pages 4581--4591, 2019.

\bibitem{wang2023internvid}
Yi~Wang, Yinan He, Yizhuo Li, Kunchang Li, Jiashuo Yu, Xin Ma, Xinyuan Chen, Yaohui Wang, Ping Luo, Ziwei Liu, Yali Wang, Limin Wang, and Yu~Qiao.
\newblock Internvid: A large-scale video-text dataset for multimodal understanding and generation.
\newblock {\em arXiv preprint arXiv:2307.06942}, 2023.

\bibitem{wang2024videohallucer}
Yuxuan Wang, Yueqian Wang, Dongyan Zhao, Cihang Xie, and Zilong Zheng.
\newblock Videohallucer: Evaluating intrinsic and extrinsic hallucinations in large video-language models.
\newblock {\em arXiv preprint arXiv:2406.16338}, 2024.

\bibitem{wu2024longvideobench}
Haoning Wu, Dongxu Li, Bei Chen, and Junnan Li.
\newblock Longvideobench: A benchmark for long-context interleaved video-language understanding.
\newblock {\em arXiv preprint arXiv:2407.15754}, 2024.

\bibitem{xiong2024llavacriticlearningevaluatemultimodal}
Tianyi Xiong, Xiyao Wang, Dong Guo, Qinghao Ye, Haoqi Fan, Quanquan Gu, Heng Huang, and Chunyuan Li.
\newblock Llava-critic: Learning to evaluate multimodal models, 2024.

\bibitem{xu2023mplug}
Haiyang Xu, Qinghao Ye, Ming Yan, Yaya Shi, Jiabo Ye, Yuanhong Xu, Chenliang Li, Bin Bi, Qi~Qian, Wei Wang, et~al.
\newblock mplug-2: A modularized multi-modal foundation model across text, image and video.
\newblock In {\em International Conference on Machine Learning}, pages 38728--38748. PMLR, 2023.

\bibitem{xu2016msr}
Jun Xu, Tao Mei, Ting Yao, and Yong Rui.
\newblock Msr-vtt: A large video description dataset for bridging video and language.
\newblock In {\em Proceedings of the IEEE conference on computer vision and pattern recognition}, pages 5288--5296, 2016.

\bibitem{xu2024pllava}
Lin Xu, Yilin Zhao, Daquan Zhou, Zhijie Lin, See~Kiong Ng, and Jiashi Feng.
\newblock Pllava: Parameter-free llava extension from images to videos for video dense captioning.
\newblock {\em arXiv preprint arXiv:2404.16994}, 2024.

\bibitem{xu2024slowfast}
Mingze Xu, Mingfei Gao, Zhe Gan, Hong-You Chen, Zhengfeng Lai, Haiming Gang, Kai Kang, and Afshin Dehghan.
\newblock Slowfast-llava: A strong training-free baseline for video large language models.
\newblock {\em arXiv preprint arXiv:2407.15841}, 2024.

\bibitem{xue2022advancing}
Hongwei Xue, Tiankai Hang, Yanhong Zeng, Yuchong Sun, Bei Liu, Huan Yang, Jianlong Fu, and Baining Guo.
\newblock Advancing high-resolution video-language representation with large-scale video transcriptions.
\newblock In {\em Proceedings of the IEEE/CVF Conference on Computer Vision and Pattern Recognition}, pages 5036--5045, 2022.

\bibitem{yan2022videococa}
Shen Yan, Tao Zhu, Zirui Wang, Yuan Cao, Mi~Zhang, Soham Ghosh, Yonghui Wu, and Jiahui Yu.
\newblock Videococa: Video-text modeling with zero-shot transfer from contrastive captioners.
\newblock {\em arXiv preprint arXiv:2212.04979}, 2022.

\bibitem{yao2012detecting}
Cong Yao, Xiang Bai, Wenyu Liu, Yi~Ma, and Zhuowen Tu.
\newblock Detecting texts of arbitrary orientations in natural images.
\newblock In {\em 2012 IEEE conference on computer vision and pattern recognition}, pages 1083--1090. IEEE, 2012.

\bibitem{yao2024minicpm}
Yuan Yao, Tianyu Yu, Ao~Zhang, Chongyi Wang, Junbo Cui, Hongji Zhu, Tianchi Cai, Haoyu Li, Weilin Zhao, Zhihui He, et~al.
\newblock Minicpm-v: A gpt-4v level mllm on your phone.
\newblock {\em arXiv preprint arXiv:2408.01800}, 2024.

\bibitem{yi2019clevrer}
Kexin Yi, Chuang Gan, Yunzhu Li, Pushmeet Kohli, Jiajun Wu, Antonio Torralba, and Joshua~B Tenenbaum.
\newblock Clevrer: Collision events for video representation and reasoning.
\newblock {\em arXiv preprint arXiv:1910.01442}, 2019.

\bibitem{zhang2023videollama}
Hang Zhang, Xin Li, and Lidong Bing.
\newblock Video-llama: An instruction-tuned audio-visual language model for video understanding.
\newblock {\em arXiv preprint arXiv:2306.02858}, 2023.

\bibitem{zhang2024eventhallusion}
Jiacheng Zhang, Yang Jiao, Shaoxiang Chen, Jingjing Chen, and Yu-Gang Jiang.
\newblock Eventhallusion: Diagnosing event hallucinations in video llms.
\newblock {\em arXiv preprint arXiv:2409.16597}, 2024.

\bibitem{zhang2024vinoground}
Jianrui Zhang, Mu~Cai, and Yong~Jae Lee.
\newblock Vinoground: Scrutinizing lmms over dense temporal reasoning with short videos.
\newblock {\em arXiv preprint arXiv:2410.02763}, 2024.

\bibitem{zhang2024internlmxcomposer}
Pan Zhang, Xiaoyi Dong, Yuhang Zang, Yuhang Cao, Rui Qian, Lin Chen, Qipeng Guo, Haodong Duan, Bin Wang, Linke Ouyang, Songyang Zhang, Wenwei Zhang, Yining Li, Yang Gao, Peng Sun, Xinyue Zhang, Wei Li, Jingwen Li, Wenhai Wang, Hang Yan, Conghui He, Xingcheng Zhang, Kai Chen, Jifeng Dai, Yu~Qiao, Dahua Lin, and Jiaqi Wang.
\newblock Internlm-xcomposer-2.5: A versatile large vision language model supporting long-contextual input and output, 2024.

\bibitem{zhang2019icdar}
Rui Zhang, Yongsheng Zhou, Qianyi Jiang, Qi~Song, Nan Li, Kai Zhou, Lei Wang, Dong Wang, Minghui Liao, Mingkun Yang, et~al.
\newblock Icdar 2019 robust reading challenge on reading chinese text on signboard.
\newblock In {\em 2019 international conference on document analysis and recognition (ICDAR)}, pages 1577--1581. IEEE, 2019.

\bibitem{zhang2024directpreferenceoptimizationvideo}
Ruohong Zhang, Liangke Gui, Zhiqing Sun, Yihao Feng, Keyang Xu, Yuanhan Zhang, Di~Fu, Chunyuan Li, Alexander Hauptmann, Yonatan Bisk, and Yiming Yang.
\newblock Direct preference optimization of video large multimodal models from language model reward, 2024.

\bibitem{zhang2024video}
Yuanhan Zhang, Jinming Wu, Wei Li, Bo~Li, Zejun Ma, Ziwei Liu, and Chunyuan Li.
\newblock Video instruction tuning with synthetic data.
\newblock {\em arXiv preprint arXiv:2410.02713}, 2024.

\bibitem{zhou2024mlvu}
Junjie Zhou, Yan Shu, Bo~Zhao, Boya Wu, Shitao Xiao, Xi~Yang, Yongping Xiong, Bo~Zhang, Tiejun Huang, and Zheng Liu.
\newblock Mlvu: A comprehensive benchmark for multi-task long video understanding.
\newblock {\em arXiv preprint arXiv:2406.04264}, 2024.

\bibitem{zhou2018youcook2}
Yipin Zhou, Zhaowen Wang, Chen Fang, Trung Bui, and Tamara~L Berg.
\newblock Visual to sound: Generating natural sound for videos in the wild.
\newblock In {\em Proceedings of the IEEE conference on computer vision and pattern recognition}, pages 3550--3558, 2018.

\bibitem{zohar2024apolloexplorationvideounderstanding}
Orr Zohar, Xiaohan Wang, Yann Dubois, Nikhil Mehta, Tong Xiao, Philippe Hansen-Estruch, Licheng Yu, Xiaofang Wang, Felix Juefei-Xu, Ning Zhang, Serena Yeung-Levy, and Xide Xia.
\newblock Apollo: An exploration of video understanding in large multimodal models, 2024.

\end{thebibliography}

\newpage
\appendix

\section{Training hyper-parameters}\label{sec:training-hyperparameters}
Table \ref{tab:hyperparam} shows the training hyper-parameters in pre-training, SFT-1\&2 and DPO stage.
We apply a layer-wise learning rate decay of 0.9 for visual encoder training \cite{chen2024internvl}. 

\begin{table}[h!]
    \centering
    \resizebox{\textwidth}{!}{
    \begin{tabular}{l |c c c c}
         \toprule
         \textbf{Configuration}   & \textbf{Pre-training} & \textbf{SFT-1} & \textbf{SFT-2} & \textbf{DPO} \\\midrule
         VLM init.                & Qwen2-VL-7B & Tarsier2-Pre-trian & Tarsier2-SFT-1 & Tarsier2-SFT-2 \\
         Optimizer name           & \multicolumn{4}{c}{AdamW} \\
         Optimizer $\beta_1$      & \multicolumn{4}{c}{$0.9$}\\
         Optimizer $\beta_2$      & \multicolumn{4}{c}{$0.999$}\\
         Optimizer eps            & \multicolumn{4}{c}{$1e^{-6}$}\\
         Learning rate            & $2e^{-5}$ & $2e^{-5}$ & $2e^{-6}$ & $1e^{-6}$\\
         Learning rate schedule   & \multicolumn{4}{c}{cosine} \\
         Training steps           & 200,000 & 5,000 & 5,000 & 1,000\\
         Warm-up steps            & 1,000 & 250 & 250 & 100 \\
         Weight decay             & \multicolumn{4}{c}{0.01}\\
         Gradient clip            & \multicolumn{4}{c}{1.0} \\
         Dropout rate             & \multicolumn{4}{c}{0.0}\\
         Global batch size        & 384 & 64 & 64 & 64 \\
         Max pixels         & \multicolumn{4}{c}{460,800} \\
         Frames per video         & [8,128] & 16 & 16 & 16 \\
         Numerical precision      & \multicolumn{4}{c}{bfloat16} \\
         \bottomrule
    \end{tabular}
    }
    \caption{Training hyper-parameters of \modelname}
    \label{tab:hyperparam}
\end{table}

\section{Public datasets of pre-training stage}\label{sec:public-datasets}

Table \ref{tab:pretraining-datasets} presents the pre-training datasets, which collectively include approximately 20 million public data and 20 million in-house data. Most of the public datasets are the same as Tarsier1, except we additionally gathered some newly released open-source data and OCR-releated data. For WebVid-10M, we used 2.9 million video-text pairs, selecting samples that are more likely to feature dynamic events. We have also incorporated some latest long video understanding datasets, such as MovieStory101\cite{he2024storyteller} and LLaVA-Video-178K~\cite{zhang2024video}. This greatly enhances the model's ability to understand long videos. 

\begin{table}[h!]
\centering
\small
\setlength{\tabcolsep}{3pt} 
\resizebox{\textwidth}{!}{
\begin{tabular}{llll}
\toprule
\multicolumn{4}{l}{\textit{\textbf{Video Captioning}}} \\
WebVid~\cite{bain2021frozen} (2.9M) & 
LSMDC~\cite{rohrbach2017movie} (109K) &
TGIF~\cite{li2016tgif} (105K) & 
ActivityNet~\cite{krishna2017dense} (38K) \\
Charades~\cite{sigurdsson2016hollywood} (16K) & 
Charades-Ego~\cite{sigurdsson2018charades} (6K) & 
YouCook2~\cite{zhou2018youcook2} (9K) &
TACoS~\cite{regneri2013grounding} (18K)\\
Ego4D~\cite{grauman2022ego4d} (1.1M) &
Spoken Moments~\cite{monfort2021spoken} (493K) & 
Multi-Moments~\cite{monfort2021multi} (997K) &
TREC-VTT~\cite{awad2023trecvid} (64K) \\
ShareGPT-4o-video~\cite{sharegpt4o} (2K) &
MovieStory101\cite{he2024storyteller} (11K)  &
GPT4o-labeled Caption$^\dagger$ (2.5M) &
Human-labeled Caption$^\dagger$ (145K) \\
Film\&TV Commentary$^\dagger$ (11.5M) & 
\\

\midrule
\multicolumn{4}{l}{\textit{\textbf{Action Recognition}}} \\
HMDB~\cite{kuehne2011hmdb} (5.8K) & 
COIN~\cite{tang2019coin} (10K) & 
SSV2~\cite{goyal2017something} (169K) &
Kinetics-700~\cite{carreira2017quo} (537K) \\
FineAction~\cite{liu2022fineaction} (82K) & 
RareAct~\cite{miech2020rareact} (2K) & 
20BN-jester~\cite{materzynska2019jester} (46K) & \\

\midrule
\multicolumn{4}{l}{\textit{\textbf{Video QA}}} \\
CLEVRER~\cite{yi2019clevrer} (83K) & 
TGIF-QA~\cite{jang2017tgif} (72K) & 
EgoQA~\cite{fan2019egovqa} (5K) &
VideoInstruct~\cite{maaz2023video} (89K) \\
LLaVA-Video-178K~\cite{zhang2024video} (165K) & 
M4-Instruct-video~\cite{li2024llava} (255K) & 
GPT4o-labeled QA$^\dagger$ (16.2K) & 
\\

\midrule
\multicolumn{4}{l}{\textit{\textbf{Grounding}}} \\
DiDeMo~\cite{anne2017localizing} (82K) & 
AVA~\cite{gu2018ava} (28K) & 
E.T. Instruct 164K~\cite{liu2024etbench} (147K) & 
Object Tracking$^\dagger$ (745K) \\

\midrule
\multicolumn{4}{l}{\textit{\textbf{Video Self-Supervised Training}}} \\
Frame Order Prediction$^\dagger$ (825K) \\

\midrule
\multicolumn{4}{l}{\textit{\textbf{Intent Recognition}}} \\
Oops!~\cite{epstein2020oops} (15K) & & & \\

\midrule
\multicolumn{4}{l}{\textit{\textbf{Multi-Image Understanding}}} \\
VIST~\cite{huang2016visual} (38K) & 
MMDU~\cite{liu2024mmdu} (45K) & 
M4-Instruct-image~\cite{li2024llava} (616K) & 
Image Retrival$^\dagger$ (533K) \\

\midrule
\multicolumn{4}{l}{\textit{\textbf{Single-Image Understanding}}} \\
ShareGPT4V~\cite{chen2023sharegpt4v} (95K) &
LLaVA-1.5~\cite{liu2023improved} (643K) &
ShareGPT-4o-image\cite{sharegpt4o} (57K) & 
MS COCO~\cite{lin2014microsoft} (566K) \\
Flicker~\cite{plummer2015flickr30k} (145K) & 
LLaVA-ReCap-CC3M~\cite{li2024llava} (2.9M) & 
Visual Genome~\cite{krishna2017visual} (759K) & 
SBU Captions~\cite{ordonez2011im2text} (860K) \\
GPT4o-labeled Caption$^\dagger$ (1.13M) \\

\midrule
\multicolumn{4}{l}{\textit{\textbf{Image OCR}}} \\
RCTW-17~\cite{shi2017icdar2017} (8K) & 
LSVT~\cite{sun2019icdar} (430K) & 
ReCTS~\cite{zhang2019icdar} (20K) & 
Art~\cite{bhagavatula2019abductive} (5.6K) \\
COCOTextV2~\cite{veit2016coco} (16K) & 
CORD-v2~\cite{park2019cord} (1K) & 
HierText~\cite{long2022towards} (10K) & 
MSRA-TD500~\cite{yao2012detecting} (465) \\
IC03~\cite{lucas2005icdar} (499) & 
SynthDoG-en~\cite{kim2022donut} (100K) & 
SynthDoG-zh~\cite{kim2022donut} (100K) & \\

\midrule
\multicolumn{4}{l}{\textit{\textbf{Text Generation}}} \\
OpenOrca~\cite{lian2023openorca} (995K) &
ShareGPT~\cite{vicuna2023} (80K) & & \\

\bottomrule
\end{tabular}
}
\caption{Datasets and their sizes used in \modelname pre-training. $\dagger$ indicates in-house datasets.}
\label{tab:pretraining-datasets}
\end{table}

\section{Annotation process for SFT data}\label{sec:SFT-prompt}
In the first stage of SFT, we annotated each video clip with detailed descriptions that included fine-grained temporal grounding.
Each clip first underwent manual annotation, where annotators described dynamic information such as character actions, events, scene transitions, and camera movements, while avoiding unnecessary static elements. Annotators are also required to map the dynamic information in their descriptions to the corresponding frame numbers. We performed quality inspections on the annotated data and returned any data not meeting quality standards for re-annotation. We discarded any data that might involve copyright risks.

In the second stage of SFT, we utilized GPT-4o to generate a variety of instruction tuning samples based on manual annotations. We provided GPT-4o with 16 uniformly sampled frames from the video and the original manual annotations. Figure \ref{fig:prompt-sft2} shows the prompt for re-annotation in this stage.

\begin{figure}[h!]
    \begin{tcolorbox}[title={The re-annotation prompt for diverse instruction data (SFT-2). }, fontupper=\scriptsize, before skip=0pt, after skip=0pt]
    \textbf{Character}
    
    You are an excellent video analyst. Utilizing your incredible attention to detail, you provide clear, sequential descriptions for video. You excel in identifying and conveying changes in actions, behaviors, environment, states and attributes of objects, and camera movements between video frames.
    
    \textbf{Prompt}

    Here are 16 frames from a video and a short video caption in Chinese. 
    You need to process a two step tasks:
    
    First, establish a set of guiding principles to control the style of the video description. These principles should include one or more of the following aspects:
    
    \begin{enumerate}[nosep, leftmargin=24pt]
        \item Specify the length constraints of the description, including the number of paragraphs and total word count.
        \item Define the level of detail for human or creature appearance, non-creature appearance, and background.
        \item Determine the granularity of the event information.
        \item Decide on the output format, such as plain text, JSON, lists, narrative, poetry, etc.
        \item Choose the output language, such as Chinese, English, Japanese, French, and so on.
        \item Decide on the text style, such as fluent, concise, professional, or just using simple words and phrases.
    \end{enumerate}
    
    Next, generate the corresponding video description based on these guiding principles and the input video clip, and rephrase the guiding principles into natural language as part of the output question.

    \textbf{Input}
    
    Origin Short Video Caption in Chinese: \{Manual Labeled Chinese Caption\}
    
    \textbf{Requirement}
    
    Return in JSON format: \{``qustion'': xxx,``answer'': xxx\}
    \end{tcolorbox}
    \vspace{5pt}
    \caption{The re-annotation prompt in SFT-2.}
    \label{fig:prompt-sft2}
\end{figure}

\section{Detail setting of DPO training}\label{sec:dpo_ab_settings}
As a default setting, we leveraged the negative sampling and preference pair filtering strategy as introduced in Section \ref{sec:dpo} to construct the DPO training set. We set top\_p as 0.7 and temperature as 0.7 when running both positive sampling and negative sampling on our 150K SFT dataset. The threshold $\delta$ of preference pair filtering was set as 0.3. We finally randomly sampled 20K preference pairs for DPO training. For the ``\textit{w/o NS}'' setting, we kept other parameters and process unchanged but replaced the negative sampling with an additional positive sampling. For the ``\textit{w/o PF}'' setting, we omitted the process of preference pair filtering and directly sample 20K pairs from all preference pair candidates. We utilized the vanilla DPO training objective (Equation \ref{eq:dpo_loss}), and set $\beta$ as 0.1. See the ``DPO'' column of Table \ref{tab:hyperparam} for all the other hyper-parameters.

\section{Detailed
results of individual datasets at different stages}\label{sec:ab_exp_detail}
In this section, we provide detailed results for individual datasets in our ablation study. Table \ref{tab:appendix-pretrain_detailed_results}, \ref{tab:appendix-sft_detailed_results} and \ref{tab:appendix-dpo_detailed_results} list the results for pre-training, SFT and DPO respectively. Table \ref{tab:appendix-recap_detailed_results} lists the results for the recaptioning experiment. We report F1/Precision/Recall for DREAM-1K and accuracy for other benchmarks.

\begin{table}[h!]
    \centering
    \resizebox{0.9\textwidth}{!}{%
    \begin{tabular}{cl|ccc}
    \toprule
        \textbf{Capability} & \textbf{Benchmark} & \textbf{Tarsier1-7B} & \textbf{Tarsier1-7B-Qwen} & \textbf{Tarsier2-7B} \\
        
    \midrule
    \multirow{3}{*}{Caption} & DREAM-1K & 34.6/30.2/40.3 & 38.4/40.6/36.4 & 40.8/42.5/39.3 \\
    & TempCompass-cg & 55.3 & 59.3 & 60.1 \\
    & Vinoground-Text & 29.8 & 48.6 & 60.2 \\
    \midrule
    \multirow{3}{*}{Video QA Short} & MVBench & 62.6 & 69.8 & 72.8  \\
    & TVBench & 45.8 & 51.0 & 53.5 \\
    & TOMATO & 28.6 & 36.5 & 39.5 \\
    \midrule
    \multirow{3}{*}{Video QA Long} & Video-MME & 42.2 & 58.9 & 65.3 \\
    & LongVideoBench & 39.8 & 52.1 & 58.3 \\
    & TemporalBench & 56.9 & 61.9 & 68.7 \\
    \midrule
    \multirow{2}{*}{Hallucination} & EventHallusion-Y/N & 70.9 & 75.6 & 77.8 \\
    & EventHallusion-Desc & 41.6 & 48.6 & 49.1\\
    \bottomrule
    \end{tabular}
    }
    \caption{Detailed results of the ablation study for pre-training. For the captioning task, results are reported after the SFT stage. For other tasks, results are reported after the pre-training stage. }
    \label{tab:appendix-pretrain_detailed_results}
\end{table}

\begin{table}[h!]
    \centering
    \resizebox{0.9\textwidth}{!}{%
    \begin{tabular}{cl|ccc}
    \toprule
        \textbf{Capability} & \textbf{Benchmark} & \makecell[c]{\\ pre-train} & \makecell[c]{\textbf{Tarsier2-7B}\\ SFT w/o grounding} & \makecell[c]{\\ SFT} \\
    \midrule
    \multirow{3}{*}{Caption} & DREAM-1K & 35.2/36.8/33.7 & 37.4/38.6/36.3  & 40.8/42.5/39.3 \\
    & TempCompass-cg & 50.5 & 50.2 & 60.1 \\
    & Vinoground-Text & 57.2 & 60.6 & 60.2 \\
    \midrule
    \multirow{3}{*}{Video QA Short} & MVBench & 72.8 & 71.9 & 72.5  \\
    & TVBench & 53.5 & 54.5 & 54.2 \\
    & TOMATO & 39.5 & 41.3 & 41.9 \\
    \midrule
    \multirow{3}{*}{Video QA Long} & Video-MME & 65.3 & 64.0 & 64.7 \\
    & LongVideoBench & 58.3 & 54.7 & 58.2 \\
    & TemporalBench & 68.7 & 66.9 & 66.6 \\
    \midrule
    \multirow{2}{*}{Hallucination} & EventHallusion-Y/N & 77.8 & 80.1 & 84.4 \\
    & EventHallusion-Desc & 49.1 & 56.2 & 59.4 \\
    \bottomrule
    \end{tabular}
    }
    \caption{Detailed results of the ablation study for SFT.}
    \label{tab:appendix-sft_detailed_results}
\end{table}

\begin{table}[h!]
    \centering
    \resizebox{0.9\textwidth}{!}{%
    \begin{tabular}{cl|cccc}
    \toprule
        \textbf{Capability} & \textbf{Benchmark} & \textbf{Tarsier2-7B} & \textit{w/o DPO} & \textit{w/o NS} & \textit{w/o PF}\\
    \midrule
    \multirow{3}{*}{Caption} & DREAM-1K & 42.0/42.8/41.1 & 40.8/42.5/39.3 & 41.5/44.5/39.0 & 40.5/39.9/41.1 \\
    & TempCompass-cg & 66.6 & 60.1 & 62.1 & 65.1 \\
    & Vinoground-Text & 65.8 & 60.2 & 60.6 & 67.6 \\
    \midrule
    \multirow{3}{*}{Video QA Short} & MVBench & 71.5 & 72.5 & 72.2 & 71.7 \\
    & TVBench & 54.7 & 54.2 & 54.9 & 54.6 \\
    & TOMATO & 42.0 & 41.9 & 41.3 & 41.8 \\
    \midrule
    \multirow{3}{*}{Video QA Long} & Video-MME & 64.5 & 64.7 & 64.3 & 64.4 \\
    & LongVideoBench & 58.6 & 58.2 & 58.6 & 57.4 \\
    & TemporalBench  & 65.3 & 66.6 & 65.4 & 65.2 \\
    \midrule
    \multirow{2}{*}{Hallucination} & EventHallusion-Y/N & 84.6 & 84.4 & 85.1 & 84.8 \\
    & EventHallusion-Desc & 63.3 & 59.4 & 60.7 & 63.5 \\
    \bottomrule
    \end{tabular}
    }
    \caption{Detailed results of the ablation study for DPO.}
    \label{tab:appendix-dpo_detailed_results}
\end{table}

\begin{table}[h!]
    \centering
    \resizebox{0.9\textwidth}{!}{%
    \begin{tabular}{cl|ccc}
    \toprule
        \textbf{Capability} & \textbf{Benchmark} & \textbf{Qwen2-VL-7B} \cite{qwen2vl} & \textit{$+$ Original FT} & \textit{$+$ Recaption FT} \\
    \midrule
    \multirow{3}{*}{Caption} & DREAM-1K & 29.6/33.9/26.3 & 35.2/44.8/29.0 & 39.5/41.7/37.6 \\
    & TempCompass-cg & 54.2 & 49.9 & 67.7 \\
    & Vinoground-Text & 40.0 & 39.0 & 55.0 \\
    \midrule
    \multirow{3}{*}{Video QA Short} & MVBench & 67.0 & 59.8 & 66.8 \\
    & TVBench & 43.8 & 47.2 & 51.1 \\
    & TOMATO & 31.5 & 33.6 & 39.5 \\
    \midrule
    \multirow{3}{*}{Video QA Long} & Video-MME & 63.3 & 56.1 & 57.0\\
    & LongVideoBench & 55.6 & 51.4 & 51.9 \\
    & TemporalBench  & 62.0 & 58.7 & 61.4 \\
    \midrule
    \multirow{2}{*}{Hallucination} & EventHallusion-Y/N & 68.6 & 39.6 & 80.7 \\
    & EventHallusion-Desc & 27.8 & 46.3 & 56.2 \\
    \bottomrule
    \end{tabular}
    }
    \caption{Detailed results of the recaptioning experiment.}
    \label{tab:appendix-recap_detailed_results}
\end{table}

\section{Tarsier2-Recap-585K Data Composition}\label{sec:recaption_composition}
Table \ref{tab:recaption_composition} lists the data composition details of Tarsier2-Recap-585K. We mainly took video caption datasets into account when picking the target datasets, together with two action recognition datasets (Kinetics-700 \cite{carreira2017quo} and SSV2 \cite{goyal2017something}), which contain video clips of durations of $5\sim10$ seconds about human actions, and a special intent recognition dataset (Oops~\cite{epstein2020oops}) to help models learn rare actions and unexpected events. For most of the datasets, we utilized all the original video clips of the selected splits (usually train and val set), except for:
\begin{itemize}
    \item WebVid-10M: We sampled around 30\% of the total size of Tarsier2-Recap-585K from a pre-filtered subset of WebVid-10M, which are more likely to feature dynamic events.
    \item Ego4D: We randomly merged multiple clips into a new one that contains multiple actions and result in around 1M merged clips in total. We sampled 50K clips from this dataset for recaptioning.
    \item Kinetics-700 and SSV2: We randomly sampled 50K and 10K clips from the training set of Kinetics-700 and SSV2, respectively.
\end{itemize}

\begin{table}[t]
    \centering
    \resizebox{\textwidth}{!}{%
    \begin{tabular}{lccrrr}
    \toprule
        \textbf{Dataset} & \textbf{Original Label Type} & \textbf{Split} & \textbf{Avg Duration (s)} & \textbf{\# Sampled Clips} & \textbf{Proportion (\%)} \\
    \midrule
    WebVid-10M~\cite{bain2021frozen} & \multirow{9}{*}{Video Caption} & - & 15.2 & 177,909 & 30.38 \\
    LSMDC~\cite{rohrbach2017movie} &  & \textbf{train}/\textbf{val}/\textbf{test} & 4.1 & 108,271 & 18.49 \\
    TGIF~\cite{li2016tgif} &  & \textbf{train}/test & 12.3 & 94,775 & 16.18 \\
    Ego4D~\cite{grauman2022ego4d} &  & - & 4.1 & 50,000 & 8.54 \\
    ActivityNet~\cite{krishna2017dense} &  & \textbf{train}/\textbf{val}/test & 35.7 & 35,960 & 6.14 \\
    VATEX~\cite{wang2019vatex} &  & \textbf{train}/\textbf{val}/test & 10.0 & 22,435 & 3.83 \\
    TREC-VTT~\cite{awad2023trecvid} & & \textbf{train}/val & 6.3 & 14,199 & 2.42 \\
    Charades~\cite{sigurdsson2016hollywood} & & \textbf{train}/test & 29.8 & 7,985 & 1.36 \\
    Charades-Ego~\cite{sigurdsson2018charades} & & \textbf{train}/test & 30.2 & 6,161 & 1.05 \\
    \midrule
    Kinetics-700~\cite{carreira2017quo} & \multirow{2}{*}{Action Recognition} & \textbf{train}/val/test & 8.9 & 50000 & 8.50 \\
    SSV2~\cite{goyal2017something} & & \textbf{train}/val/test & 3.7 & 10000 & 1.71 \\
    \midrule
    Oops~\cite{epstein2020oops} & Intent Recognition & \textbf{train}/\textbf{val} & 9.8 & 7,948 & 1.36 \\
    \midrule
    \textbf{Sum} & - & - & \textbf{1,972 hours} & \textbf{585,643} & \textbf{100.00} \\
    \bottomrule
    \end{tabular}
    }
    \caption{Data composition of Tarsier2-Recap-585K. The ``Split'' column lists the original dataset partitioning, and we use bold to mark the parts which we sampled the video clips from to conduct recaptioning.}
    \label{tab:recaption_composition}
\end{table}


\begin{figure}[h!]
    \centering
    \includegraphics[width=\linewidth]{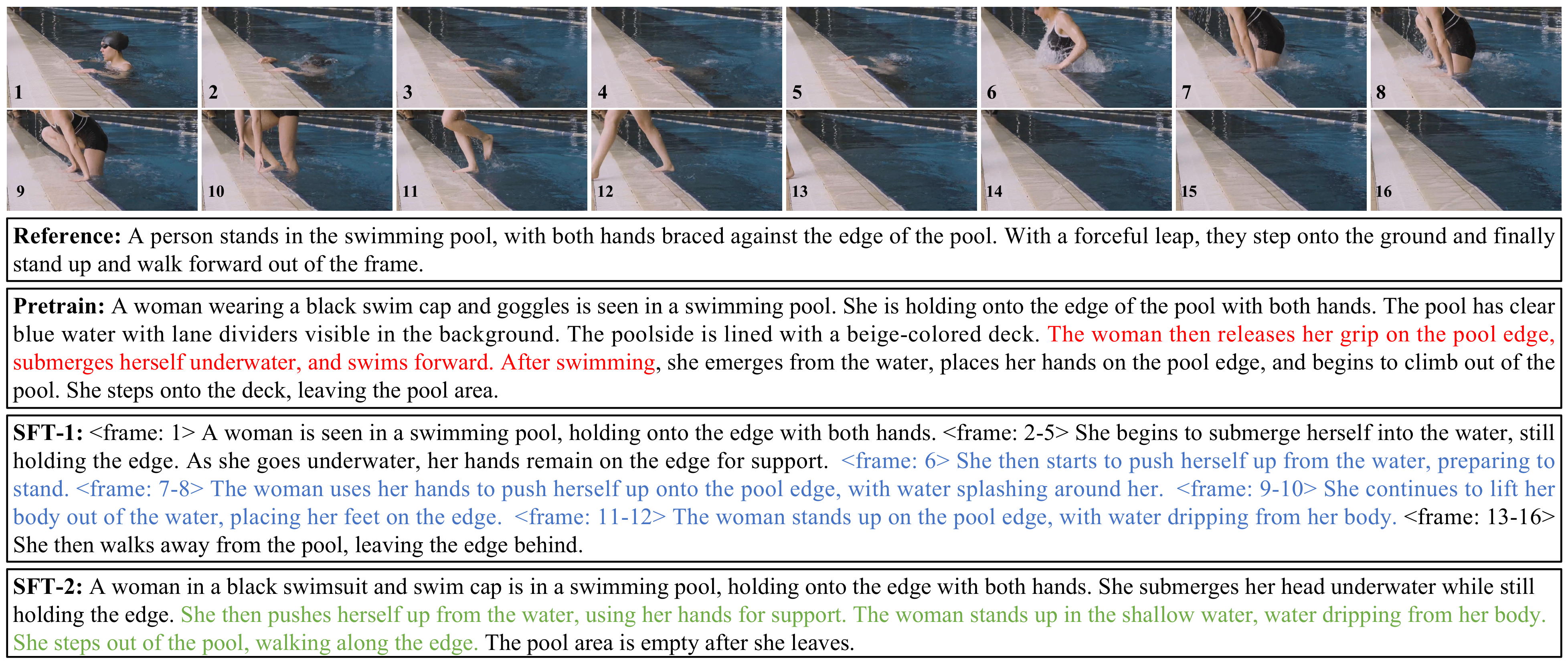}
    \caption{Qualitative comparison of our model at different stages.}
    \label{fig:sft}
\end{figure}

\section{Qualitative Comparison of the SFT Process}
Figure \ref{fig:sft} illustrates a qualitative comparison of our model at different stages, where we mark the differences in the prediction results of different models. From these differences, it can be seen that introducing temporal localization information in the first SFT stage significantly reduces the problem of hallucination in the model. However, the introduction of temporal localization information may also result in certain events being subdivided into finer actions. To address this issue, the second stage of training further improved the accuracy of the model description and optimized the output style.


\section{DREAM-1K cases}\label{sec:dream_cases}
\begin{figure}[h!]
    \centering
    \begin{tikzpicture}
        \node [anchor=south west,inner sep=0] (image) at (0,0) {\includegraphics[width=\linewidth]{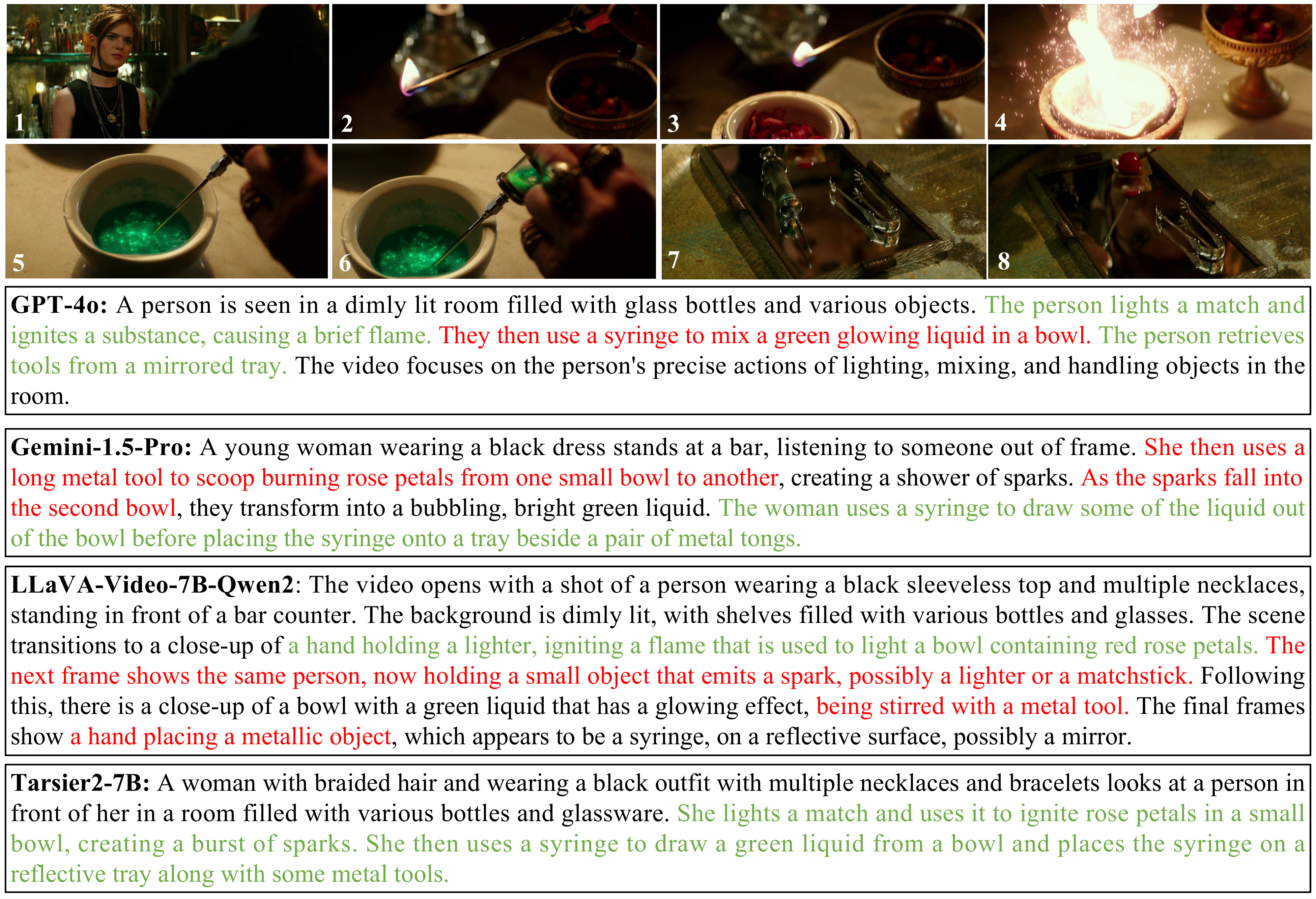}};
        \begin{scope}[x={(image.south east)},y={(image.north west)}]
            \node[anchor=north west] at (0.1,0.96) {\href{https://dream-videos.s3.us-east-1.amazonaws.com/230.mp4}{\includegraphics[width=0.04\textwidth]{figs/play.png}}};
        \end{scope}
    \end{tikzpicture}
    \caption{Qualitative comparative analysis of various Video-MLLMs on Dream-1K dataset (Live-action Subset).}
    \label{fig:Live-action}
\end{figure}

\begin{figure}[h!]
    \centering
    \begin{tikzpicture}
        \node [anchor=south west,inner sep=0] (image) at (0,0) {\includegraphics[width=\linewidth]{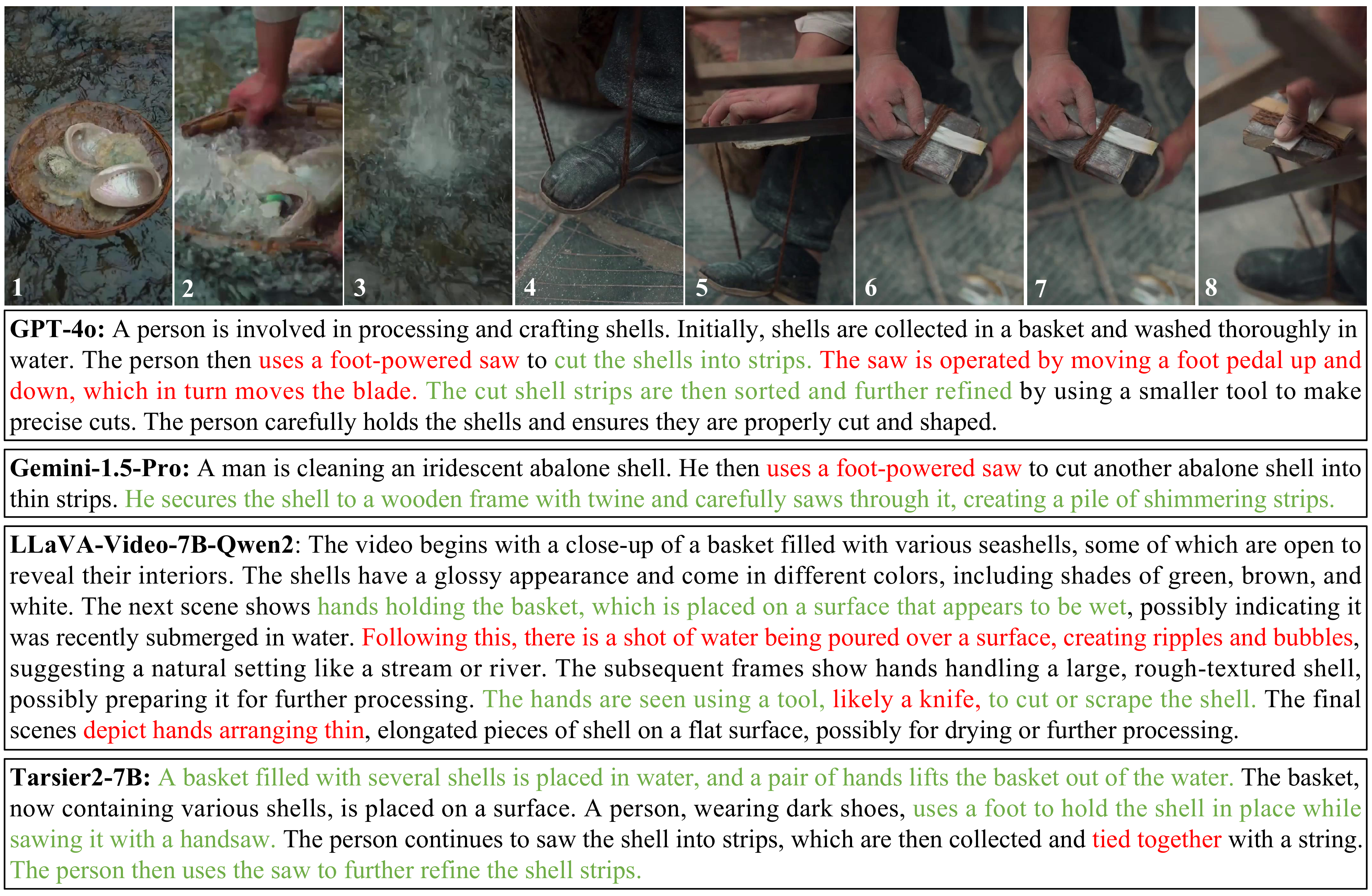}};
        \begin{scope}[x={(image.south east)},y={(image.north west)}]
            \node[anchor=north west] at (0.035,0.85) {\href{https://dream-videos.s3.us-east-1.amazonaws.com/433.mp4}{\includegraphics[width=0.04\textwidth]{figs/play.png}}};
        \end{scope}
    \end{tikzpicture}
    \caption{Qualitative comparison of different Video-MLLMs on Dream-1K dataset (Shorts Subset).}
    \label{fig:Shorts}
\end{figure}

Figure \ref{fig:Live-action}$\sim$\ref{fig:Shorts} display the detailed video descriptions generated by Tarsier2-7B and other models (GPT-4o, Gemini-1.5-Pro and LLaVA-Video-7B-Qwen2) for different video categories in DREAM-1K. Click the play button on the first frames to view the raw video. The correct descriptions of key objects/actions/events are marked in green, and the incorrect descriptions are marked in red.

\begin{figure}[h!]
    \centering
    \begin{tikzpicture}
        \node [anchor=south west,inner sep=0] (image) at (0,0) {\includegraphics[width=\linewidth]{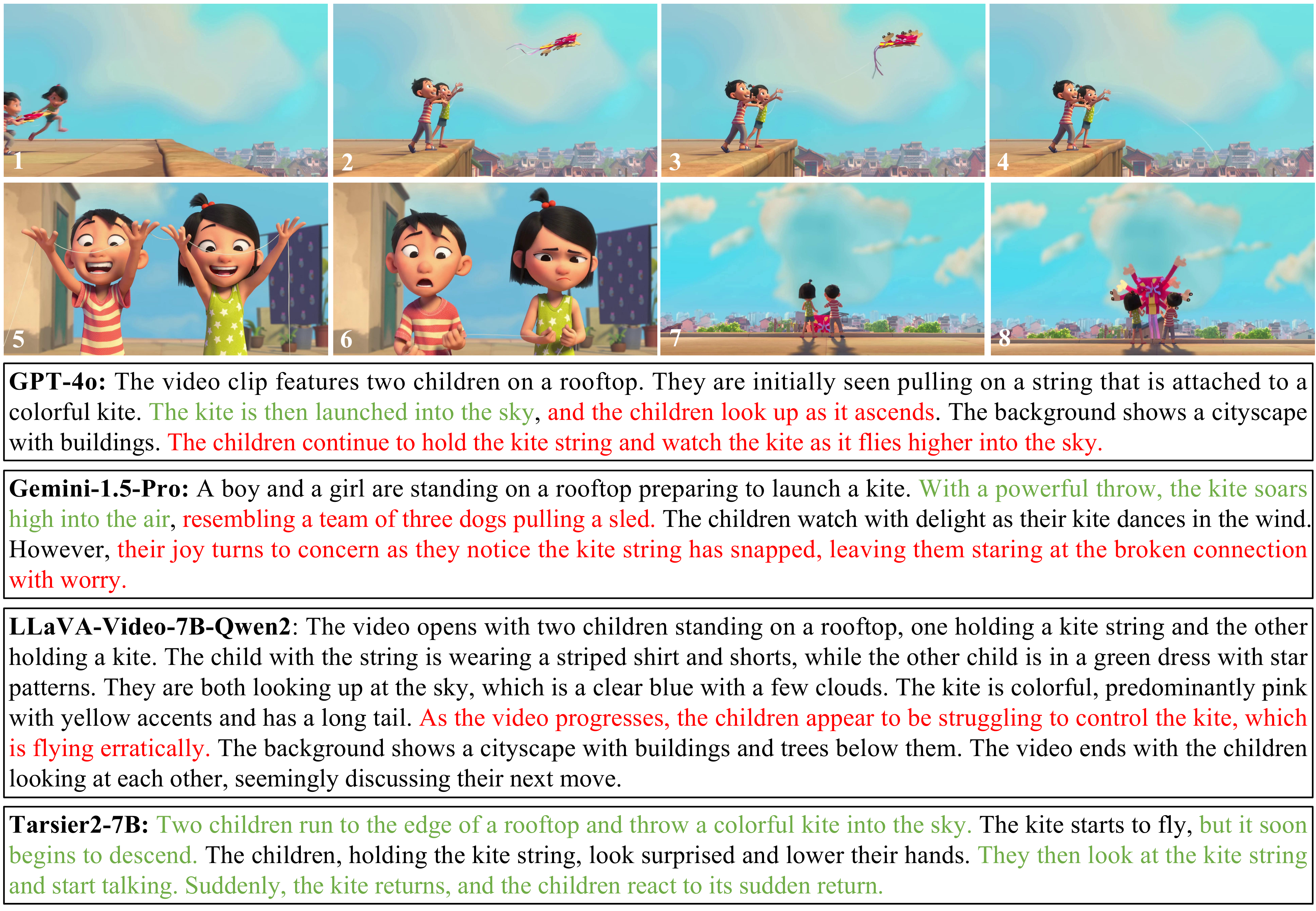}};
        \begin{scope}[x={(image.south east)},y={(image.north west)}]
            \node[anchor=north west] at (0.1,0.94) {\href{https://dream-videos.s3.us-east-1.amazonaws.com/80.mp4}{\includegraphics[width=0.04\textwidth]{figs/play.png}}};
        \end{scope}
    \end{tikzpicture}
    \caption{Qualitative comparative analysis of various Video-MLLMs on Dream-1K dataset (Animation Subset).}
    \label{fig:Animation}
\end{figure}

\begin{figure}[h!]
    \centering
    \begin{tikzpicture}
        \node [anchor=south west,inner sep=0] (image) at (0,0) {\includegraphics[width=\linewidth]{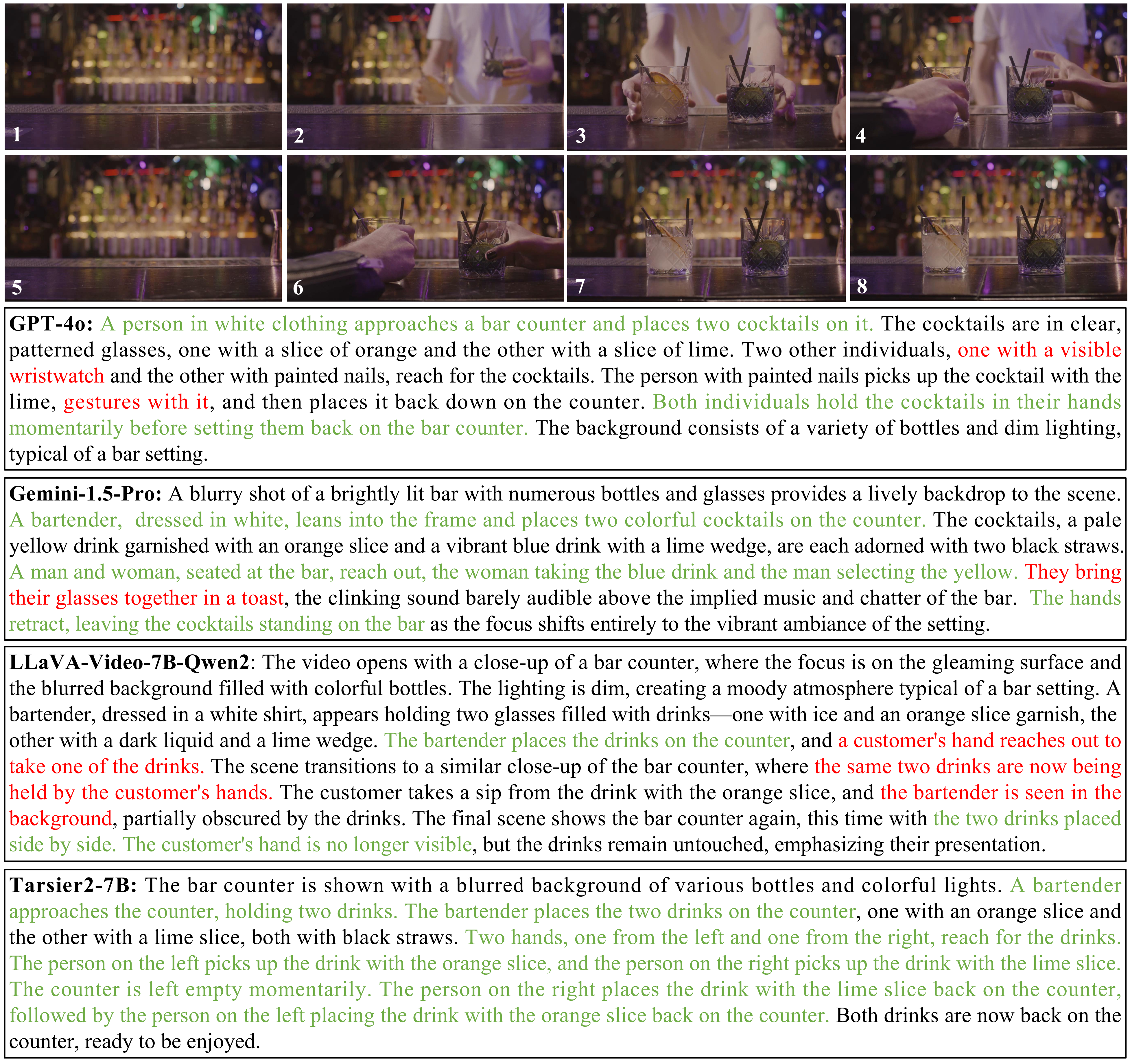}};
        \begin{scope}[x={(image.south east)},y={(image.north west)}]
            \node[anchor=north west] at (0.1,0.96) {\href{https://dream-videos.s3.us-east-1.amazonaws.com/603.mp4}{\includegraphics[width=0.04\textwidth]{figs/play.png}}};
        \end{scope}
    \end{tikzpicture}
    \caption{Qualitative comparative analysis of various Video-MLLMs on Dream-1K dataset (Stock Subset).}
    \label{fig:Stock}
\end{figure}

\begin{figure}[h!]
    \centering
    \begin{tikzpicture}
        \node [anchor=south west,inner sep=0] (image) at (0,0) {\includegraphics[width=\linewidth]{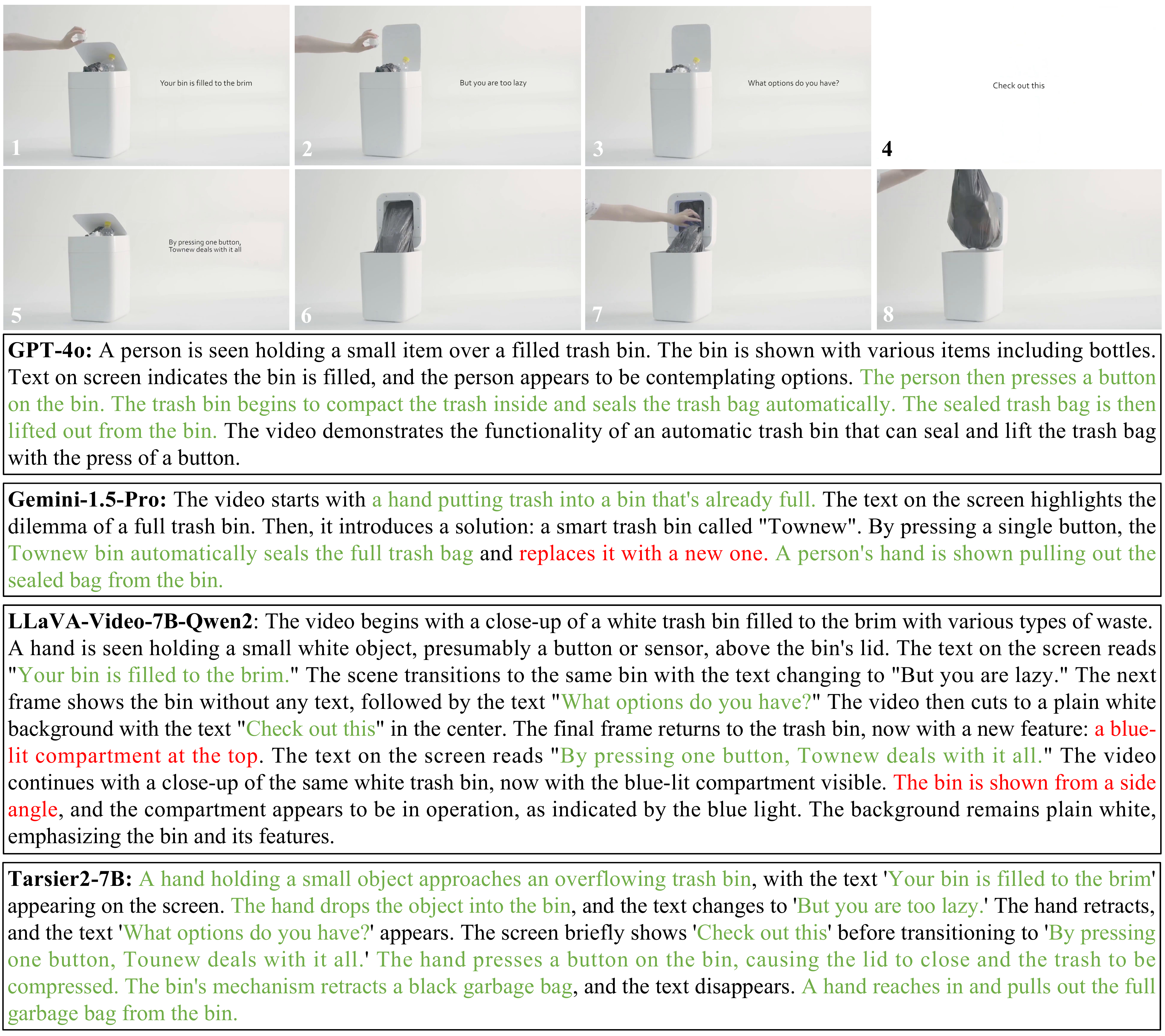}};
        \begin{scope}[x={(image.south east)},y={(image.north west)}]
            \node[anchor=north west] at (0.1,0.94) {\href{https://dream-videos.s3.us-east-1.amazonaws.com/922.mp4}{\includegraphics[width=0.04\textwidth]{figs/play.png}}};
        \end{scope}
    \end{tikzpicture}
    \caption{Qualitative comparative analysis of various Video-MLLMs on Dream-1K dataset (Youtube Subset).}
    \label{fig:Youtube}
\end{figure}

\end{document}